\pdfoutput=1

\documentclass[11pt]{article}

\usepackage[preprint]{coling}

\usepackage{times}
\usepackage{latexsym}

\usepackage{linguex}
\usepackage[greek,english]{babel}

\usepackage{url}

\usepackage{tipa}
\newcommand{\ipa}[1]{\textipa{#1}}\usepackage[utf8]{inputenc}

\usepackage{microtype}

\usepackage{inconsolata}

\usepackage{graphicx}

\newcommand\blfootnote[1]{%
  \begingroup
  \renewcommand\thefootnote{}\footnote{#1}%
  \addtocounter{footnote}{-1}%
  \endgroup
}

\title{Why do language models perform worse for morphologically complex languages?}

\author{Catherine Arnett \\
  Department of Linguistics \\
  University of California San Diego\\
  \texttt{ccarnett@ucsd.edu} \\\And
  Benjamin K. Bergen \\
  Department of Cognitive Science \\
  University of California San Diego \\
  \texttt{bkbergen@ucsd.edu} \\}

\begin{document}
\maketitle
\begin{abstract}

Language models perform differently across languages. It has been previously suggested that morphological typology may explain some of this variability \citep{cotterell-etal-2018-languages}. We replicate previous analyses and find additional new evidence for a performance gap between agglutinative and fusional languages, where fusional languages, such as English, tend to have better language modeling performance than morphologically more complex languages like Turkish. We then propose and test three possible causes for this performance gap: morphological alignment of tokenizers, tokenization quality, and disparities in dataset sizes and measurement. To test the morphological alignment hypothesis, we present MorphScore, a tokenizer evaluation metric, and supporting datasets for 22 languages. We find some evidence that  tokenization quality explains the performance gap, but none for the role of morphological alignment. Instead we find that the performance gap is most reduced when training datasets are of equivalent size across language types, but only when scaled according to the so-called ``byte-premium"---the different encoding efficiencies of different languages and orthographies. 
These results suggest that no language is harder or easier for a language model to learn on the basis of its morphological typology. Differences in performance can be attributed to disparities in dataset size. These results bear on ongoing efforts to improve performance for low-performing and under-resourced languages. 

\end{abstract}

\section{Introduction}

An enduring goal in NLP is to develop language-general systems\blfootnote{All code and data for this paper available below.  \\\url{https://osf.io/jukzd/?view_only=3d0d491d24074215a0ab81f72a693c16}} that achieve equal performance on all languages \citep{bender2011achieving}. Yet to date performance on languages other than English and a small number of high-resource languages remains extremely poor \citep{joshi-etal-2020-state, ranathunga-de-silva-2022-languages, sogaard-2022-ban, atari2023humans, ramesh-etal-2023-fairness}. This has been attributed to a lack of research on non-English languages \citep{blasi-etal-2022-systematic}, a lack of training data, and the possibility that evaluations are skewed towards high-resource languages \citep{choudhury2023generative}.

Beyond these systemic biases, it's also possible that certain linguistic features lead to higher or lower language modeling performance. Specifically, it has been proposed that languages with more complex morphology are harder to model \citep{cotterell-etal-2018-languages, park2021morphology}. Languages with more inflectional classes are morphologically more complex, and thus harder to predict. This can be described in terms of enumerative complexity \citep{ackerman2013morphological}. 

Greater morphological complexity may lead to worse language model performance, as morphologically rich languages tend to have a large number of very infrequent word forms produced by combinations of morphemes, which leads to data sparsity \citep{shin2009hybrid, bender2011achieving, botev-etal-2022-deciphering}.
This claim finds empirical support in \citet{gerz2018language}, who demonstrated over a sample of 50 languages that morphologically rich (agglutinative) languages performed worse than less morphologically rich (fusional) languages. In the current work (\S\ref{sec:perf_gap}), we replicate this analysis and extend it to much larger transformer models, both in monolingual and multilingual settings. We, too, find a robust performance gap between agglutinative and fusional languages. 

This effect is surprising, as there are reasons to think that agglutinating languages should be \textit{easier} for language models to learn.  
In studies on first language acquisition, children are observed to acquire more complex morphological systems earlier, especially systems that are uniform and transparent \citep{dressler2010typological}. This may be due to the fact that the form-meaning correspondences in these systems are more transparent, and thus more informative \citep{slobin1973cognitive, slobin2013crosslinguistic, slobin200114, dressler2010typological}. 
By adulthood, there are no observed cross-linguistic differences in the level of acquisition of different languages according to morphological typology. Therefore, there is no linguistic evidence that would predict that any language should be harder to learn than any other language. 

Identifying the causes for this performance gap could permit improved performance for morphologically rich languages (which are often low-resource) and reduce the performance inequity, potentially enabling users and researchers to be better able to use and do research on language models \citep{khanuja-etal-2023-evaluating} in their own languages. 
We evaluate three possible explanations. 

\subsection*{Hypothesis 1: Tokenization is not Morphologically Aligned}

When the token boundaries for a given word line up with its morpheme boundaries, that tokenization is morphologically aligned. For example, the word `books' in English is composed of the root `book' and the plural morpheme `-s'. A morphologically aligned tokenization would be [`book', `s']. By contrast, ['boo', 'ks'] and ['b', ooks'] would be morphologically misaligned tokenizations.

Morphological alignment of the tokenizer -- or lack thereof -- could impact language modeling performance, especially for morphologically rich languages. For these languages, relatively frequent morphemes are combined to create a large number of unique word forms, which may be rare or completely novel. If the tokenizer does not segment words along morphological boundaries, it may be difficult for the language model to efficiently learn and represent the structure of the language. Additionally, this may be further exaggerated for morphologically complex languages, which tend to have longer words.

This hypothesis would predict that agglutinative languages have less morphologically aligned tokenizers than fusional languages and that morphological alignment negatively correlates with metrics of language model performance. 
To test this hypothesis, Section \ref{sec:morphscore} introduces \textbf{MorphScore}, tokenizer evaluation for morphological alignment in 22 languages. To our knowledge, this is the first such multilingual evaluation for morphological alignment of tokenizers. 

\subsection*{Hypothesis 2: Tokenization is Worse}

Second, morphologically rich languages might tend to engender lower quality tokenizations. There is no current consensus on how to evaluate intrinsic tokenization quality \citep{zouhar-etal-2023-tokenization, chizhov2024picky}. But compression is one of the most widely used metrics (\citealp[\textit{inter alia}]{galle-2019-investigating, rust-etal-2021-good}). It is usually measured as sequence length -- the number of tokens needed to encode a sequence -- or corpus token count (CTC; \citealp[]{schmidt2024tokenization}).
Better compression has been linked to better language modeling performance because it allows for more language data to fit into a fixed sequence length \citep{galle-2019-investigating, liang-etal-2023-xlm, dagan2024getting, goldman2024unpacking}; however, there is some evidence to suggest that compression is not directly linked to performance \citep{deletang2024language, schmidt2024tokenization}. 

Agglutinative languages might have worse compression on average because words tend to be longer \citep{fenk1999cognition, berg2022effects} and there are more unique word forms \citep{sandra1994morphology}. In Turkish, for example, a single root may have millions of unique word forms \citep{hakkani2002statistical}. It is therefore less likely that the tokenizer will store whole words in its vocabulary, instead representing words using multiple tokens. This in turn may lead to worse compression and thus worse performance. If we find worse compression for agglutinative languages than for fusional languages, this may indicate that suboptimal compression is related to the performance gap. 

Another proposed metric of tokenization quality is Rényi entropy \citep{zouhar-etal-2023-tokenization}, which measures how evenly distributed token frequencies are over the whole vocabulary, penalizing very high- and very low-frequency tokens. Rényi entropy has been shown to be predictive of downstream task performance (ibid). Because of their larger number of low-frequency word forms, it is possible that agglutinative languages have higher numbers of low-frequency tokens (specific inflectional forms) or higher numbers of high-frequency tokens (very high frequency morphemes used in many different word forms) than fusional languages. Therefore if agglutinative languages display worse (higher) Rényi entropy than fusional languages, this could indicate that inefficient token frequency distribution contributes to the performance gap.

In Section \ref{sec:rq2}, we collect both compression and Rényi entropy and test whether agglutinating languages have worse compression and Rényi entropy, which would suggest that aspects of tokenization quality are driving the performance gap.

\subsection*{Hypothesis 3: Less Training Data}

The role of data quantity for pre-trained language models is uncontroversial: the more, the better. In some cases, increasing data can improve performance more than increasing model size \citep{hoffmann2022training}. Western European high-resource languages tend to be less morphologically rich, and correspondingly, many morphologically rich languages are low-resource. Morphologically rich languages have less annotated data \citep{botev-etal-2022-deciphering} and are less well researched. According to a survey by \citeauthor{blasi-etal-2022-systematic}, despite having more speakers than most European languages, morphologically complex languages like Bengali, Swahili, and Korean have only a small number of studies. German, Romanian, French, and Italian have been better studied, despite having many fewer speakers (\citealp[Table 2]{blasi-etal-2022-systematic}). Therefore, data scarcity may be driving the observed performance gap between agglutinative languages and fusional languages.

Furthermore, recent work has shown that there are disparities in the number of bytes needed to convey the same amount of information in different languages (\textit{byte premium}; \citealp[]{arnett-etal-2024-bit}), due to orthographic encoding and linguistic reasons. Morphologically rich languages are more often written with non-Latin scripts, which require more bytes to be represented in common encoding standards like UTF-8. Morphologically rich languages also have longer words, which may amplify the effect. Byte premiums may thus exacerbate the data scarcity problem, and agglutinative languages may be trained on effectively less data even than it currently seems. 
Section \ref{sec:rq3} asks whether monolingual language models trained on byte-premium-scaled text demonstrate the previously observed performance gap.

\section{Background}

\subsection{Morphological Typology}
The field of morphological typology seeks to categorize languages according to their word formation strategies \citep{brown2010morphological}. Some languages primarily use words composed of a single morpheme or a small number of morphemes. Other languages incorporate many morphemes into a single word. This paper focuses on two types of languages: fusional and agglutinative languages. 
Fusional languages tend to encode multiple morpho-syntactic features into a single morpheme, where agglutinative languages tend to use different morphemes to represent each feature \citep{plank1999split, haspelmath2009empirical, dressler2010typological}. As a result, agglutinative languages also tend to be polysynthetic (having words composed of many individual morphemes; \citealp[]{baker1996polysynthesis}). For example, Turkish has separate plural and accusative morphemes, but in English, the root, tense, number, and person may all be loaded onto a single morpheme (Exs. \ref{type_agg} and \ref{type_fus}). 

\ex.  tarla-lar-ı  \hfill (Turkish)\label{type_agg}  \\
field-PL-ACC \\
\citep{plank1999split}

\ex. are  \\
be-\textsc{PRES}.\textsc{2PL}  \hfill (English) \label{type_fus} 

Typological categorization is much more complex than this binary categorical distinction. In order to connect this work with previous studies, it is helpful to use a very coarse view of morphological type; however, these properties are gradient. Languages may have both fusional and agglutinative properties (and properties of other morphological types, too). See \citeauthor{plank1991abundance} (\citeyear{plank1991abundance}; pp. 11-16) for discussion on this point. 

\subsection{Morphologically Aligned Tokenization}

There is an area of active research on the relationship between morphological alignment of tokenizers and how it relates to language model performance. Work in this area often stems from the assumption that morphologically aligned tokenization is the gold standard for tokenization (\citealp[\textit{inter alia}]{hofmann-etal-2022-embarrassingly, bauwens-delobelle-2024-bpe, libovicky2024lexically}). Morphologically-aware or aligned tokenization has been argued to lead to more meaningful tokens, which in turn leads to better performance \citep{banerjee-bhattacharyya-2018-meaningless, klein-tsarfaty-2020-getting, tan-etal-2020-mind, hofmann-etal-2021-superbizarre, hofmann-etal-2022-embarrassingly, minixhofer-etal-2023-compoundpiece, bauwens-delobelle-2024-bpe}.
There is empirical evidence from several languages to support this claim, e.g. English \citep{jabbar2023morphpiece}, Korean \citep{lee-etal-2024-length} , Latvian \citep{pinnis2017neural}, Arabic \citep{tawfik-etal-2019-morphology}, Japanese \citep{bostrom-durrett-2020-byte}, Hebrew \citep{gueta2023explicit}, Kinyarwanda \citep{nzeyimana-niyongabo-rubungo-2022-kinyabert}, and Uyghur \citep{abulimiti-schultz-2020-building}.
However, these efforts are limited by the availability of morphologically annotated datasets \citep{minixhofer-etal-2023-compoundpiece}, which are often only available for a small number of relatively high-resource languages. 

Some evidence also exists to the contrary \citep{zhou2018morphological, minixhofer-etal-2023-compoundpiece, gutierrez2023languages}, even for some of the same languages. 
Work on German and Czech \citep{machavcek2018morphological}, Nepali, Sinhala, and Kazakh \citep{saleva-lignos-2021-effectiveness}, Korean \citep{choo2023study}, Turkish \citep{kaya2024effect}, and Spanish \citep{arnett-etal-2024-different} did not show any benefit of morphologically aligned tokenization. This is consistent with other work, e.g. \citet{uzan2024greed}, showing that BPE, which generally performs best on metrics such as compression, has the least morphologically meaningful tokens compared to other tokenization algorithms. 

\section{Evidence for a Performance Gap} \label{sec:perf_gap}

This section describes three analyses that show lower performance for agglutinative languages. Previous analyses, which demonstrated evidence for the performance gap between fusional and agglutinative languages, all had significant confounds. We extend previous work by additionally controlling for amount of training data and extending to models which--as they are much larger and use the transformer architecture--better represent the state of the field. 

\subsection{Reanalysis of \citet{gerz2018language}} \label{sec:gerz_reanalysis}

\citet{gerz-etal-2018-relation} analyzed a multilingual LSTM trained on 50 languages and found that fusional languages categorically outperformed agglutinative languages. This seminal finding is nevertheless limited in ways. It did not control for the number of training tokens, which was different for each language. We addressed this in a replication of the analysis on the original data, fitting a full linear model in R with morphological type and number of training tokens as fixed effects, predicting perplexity. We fit a reduced model with only number of training tokens as a fixed effect. An ANOVA showed that the full model explained more variance in the data than the reduced model (F(3, 45)=5.221, p=0.004)).  After controlling for number of training tokens, there is still a significant effect of morphological type, where agglutinative languages had higher perplexities than fusional languages.

\subsection{Multilingual Models} \label{sec:multilingual_analysis}

Evidence from \citet{gerz2018language} comes from just one model. To extend this work, we test a number of more contemporary multilingual language models, including XGLM \citep{lin-etal-2022-shot}, BLOOM \citep{le2023bloom}, mT0 \citep{muennighoff-etal-2023-crosslingual}, MaLA \citep{lin2024mala}, and LLaMA2 \citep{touvron2023llama}. 

We test these models across a variety of benchmarks: commonsense reasoning benchmark scores from XStoryCloze \citep{lin-etal-2022-shot}, XCOPA \citep{ponti-etal-2020-xcopa}, XNLI \citep{conneau2018xnli}, Wikipedia \citep{guo-etal-2020-wiki}, and XWinograd \citep{muennighoff-etal-2023-crosslingual} reported in the BigScience BLOOM evaluation results\footnote{\url{https://huggingface.co/datasets/bigscience/evaluation-results}} and the
SIB-200 benchmark \citep{adelani-etal-2024-sib}, as reported in the release paper. 

We combine all of the benchmark scores into one dataset. All scores are on a scale between 0 and 1. We use language family information from the WALS database \citep{dryer2013wals} and annotate the morphological type according to grammars and linguistic articles about each language. For each language model, we calculate the proportion of training data for each language according to reported data quantities in tokens or bytes. If languages were upsampled for model training, we include upsampled proportions. 

We fit a full linear mixed effects model in R \citep{bates2010lme4} predicting benchmark score with morphological type and language family as fixed effects and model and task as random effects. We fit a reduced model that is the same as the full model, except without morphological type as a predictor. We run an ANOVA to compare model fit. We find that the full model (with morphological type as a fixed effect) explains more variance than the reduced model ($\chi^{2}(3)=149.16$, p$<$0.001).
Even after controlling for amount of training data, language family, model, and benchmark task, there is still a significant effect of morphological type, where fusional languages show better performance than agglutinative languages. 

\subsection{Monolingual Models} \label{sec:monolingual_evidence}

Both of the previous analyses measure performance of multilingual models. None of these models had controlled or balanced amounts of training data for the languages they were trained on. This introduces a confound, because European languages are typically both higher resource and fusional. The lower-resource languages in this sample were more likely to be agglutinative. In this final analysis of the performance gap, we compare performance of a suite of 1,989 monolingual models from \citet{chang2023multilinguality}, covering 252 languages, which were trained on matching numbers of tokens. For each language, there are up to 12 models, with up to three different model sizes and four different training corpus sizes. 
The three model sizes were tiny (4.6M parameters), mini (11.6M parameters), and small (29.5M parameters). The four dataset sizes were low-resource (1M tokens), medlow-resource (10M tokens), medhigh-resource (100M tokens), and high-resource (1B tokens). Perplexities were calculated using 500k held-out tokens. 
We use the same language family and morphological type data as in \S\ref{sec:multilingual_analysis}. 

We use perplexity as a metric of performance. This is the only existing evaluation metric for all the languages represented by these models. 

We fit a full linear regression with morphological type, model size, and dataset size as predictors. We also fit a reduced model with only model size and dataset size as predictors. We use an ANOVA to compare the fit of these two models. We find that morphological type explains variance above and beyond the other two predictors ($\chi^{2}(3)=28.809$, p$<$0.001). We also fit full and reduced models with language family as an additional predictor. Even after accounting for language family, morphological type still explains additional variance ($\chi^{2}(3)=3.3324$, p=0.02).

Morphological type is predictive of performance after controlling for model size and data amounts, which supports the other analyses. 

\subsection{Interim Discussion}

Using both perplexities and benchmark scores as evaluation metrics, and evaluations from monolingual and multilingual models, we found a robust performance gap between agglutinative languages and fusional languages. This evidence amplifies prior work by using more evaluation metrics for more languages, with more contemporary multilingual and monolingual models trained with balanced training data. 

The following sections test three factors that may be driving this gap corresponding to the three hypotheses above: morphological alignment of the tokenizers (\S\ref{sec:rq1}), tokenization quality (\S\ref{sec:rq2}), and disparities in data measurement (\S\ref{sec:rq3}).

\section{H1: Morphological Alignment} \label{sec:rq1}

Does differential morphological alignment of tokenizers in languages with more or less complex morphology explain their performance gap? We present a new evaluation framework, called MorphScore, which permits a comparison of morphological alignment across tokenizers and languages. We evaluate monolingual tokenizers for 22 languages and analyze the relationship between MorphScore and morphological type. Code and datasets for MorphScore are available on GitHub: \url{https://github.com/catherinearnett/morphscore}.

\subsection{MorphScore: Evaluating Morphological Alignment of Tokenizers} \label{sec:morphscore}

\paragraph{Calculating MorphScore.} To evaluate a tokenizer's MorphScore for each word in a test set, we assign a value of \texttt{1} if the tokenizer places a token boundary at the morpheme boundary of interest, regardless of other token boundaries. We assign a value of \texttt{0} if there is not a token boundary at the morpheme boundary of interest.  We exclude items which contain no token boundaries (i.e. the entire word form is in the tokenizer's vocabulary), so as not to penalize the tokenizer for not segmenting the word. MorphScore is the mean of the assigned values across the dataset for a given language. See Table \ref{tab:morphscore_example} for examples.

\paragraph{Languages.} MorphScore uses datasets of morphologically annotated words. We created datasets for 22 languages, which are listed in Appendix \ref{app:morphscore}. Half are agglutinative languages and half are fusional, according to grammars and descriptions of the languages. Language selection was also balanced for resource level, where about half of the languages of each morphological type are higher-resource, and half lower-resource. The sample was designed to be as diverse as possible in terms of language family and writing system, given the other constraints. Note that all fusional languages in the sample are Indo-European, which reflects the distribution of fusional languages in the world's languages, but not all Indo-European languages are fusional (e.g. Armenian). Among the Indo-European languages, two are from the Indic branch (Gujarati, Urdu) and a variety of subgroups are represented: Slavic (Bulgarian, Slovenian, Croatian), Baltic (Lithuanian), Hellenic (Greek), Armenian, Germanic (Swiss German, Icelandic), and Celtic (Irish). The other language families represented in the sample are Japonic, Koreanic, Dravidian, Kartvelian, Austronesian, Turkic, Niger-Congo, and Uralic, as well as an isolate (Basque).

\paragraph{Datasets.} Each dataset is composed of words with their morpheme boundary annotations from Universal Dependencies\footnote{\url{https://universaldependencies.org/}} (UD) or UniMorph\footnote{\url{https://unimorph.github.io/}}. Words in MorphScore do not contain any umlaut or suppletion and 
the whole word form can be composed of the lemma and the morpheme (or the two morphemic units annotated). Most of the datasets only had one morpheme boundary annotation per word, with the exception of the Korean datasets. For Korean, when multiple morpheme boundaries were annotated, we chose the left-most boundary. We deduplicated items, and chose a random sample of 2000 for sets where there were more than 2000 items. We only included languages with at least 100 items. 

\begin{table*}[h!]
  \centering
  \begin{tabular}{lllll}
    \hline
\textbf{Language} & \textbf{Word} & \textbf{Source} & \textbf{Segmentation}              & \textbf{Score} \\ \hline
\textbf{Basque}            & aldiz         & morphemic            & aldi + z                   &                \\
                  &               & Tokenizer 1     & {[}`al', `diz'{]}          & 0            \\
                  &               & Tokenizer 2     & {[}`aldi', `z'{]}          & 1            \\\hline
\textbf{Croatian}          & suučesnika    & morphemic            & suučesnik + a                      &                \\
                  &               & Tokenizer 1     & {[}`su', `uče', `s', `nika'{]}     & 0            \\
                  &               & Tokenizer 2     & {[}`su', `u', `če', `s', `nika'{]} & 0           \\ \hline
\textbf{Icelandic }        & samrá\ipa{\dh}s       & morphemic            & samrá\ipa{\dh} + s                         &                \\
                  &               & Tokenizer 1     & {[}`samrá\ipa{\dh}', `s'{]}                & 1            \\
                  &               & Tokenizer 2     & {[}`samrá\ipa{\dh}s'{]}                    & exclude    \\   \hline
\textbf{Greek} & \textgreek{Αδριανής}& morphemic & \textgreek{Αδριανή}  + \textgreek{ς}	&  \\
& & Tokenizer 1 & [`\textgreek{Α}', `\textgreek{δ}', `\textgreek{ριανής}'] & 0 \\
& & Tokenizer 2 & [`\textgreek{Α}', `\textgreek{δρ}', `\textgreek{ιανή}', `\textgreek{ς}'] & 1 \\ \hline

  \end{tabular}
  \caption{Example items with morphemic segmentations and tokenizations with MorphScores according to their morphological alignment.}
  
  \label{tab:morphscore_example}
\end{table*}

\subsection{Tokenizers}

We use the monolingual tokenizers from \citet{chang2023multilinguality}, which are from the same models used for perplexities in \S\ref{sec:monolingual_evidence}. Each tokenizer is a SentencePiece \citep{kudo-richardson-2018-sentencepiece} tokenizer with a vocabulary size of up to 32k. Each tokenizer is trained on 10k lines of text randomly sampled from the model training data. 

\subsection{Results}

We evaluate the tokenizers on their corresponding MorphScore dataset. 
MorphScores are reported in full in Table \ref{tab:morphscore_com} in Appendix \ref{app:morphscore}. 
In order to address Hypothesis 1, we first conduct a two sample \textit{t}-test to evaluate whether agglutinative languages have lower MorphScores than fusional languages. This would be consistent with the explanation that tokenizers are more likely to fail to align token boundaries with morpheme boundaries in agglutinative languages. To the contrary, we find that agglutinative languages have higher MorphScores (M=66.3\%) than fusional languages (M=53.3\%), a significant difference (t(20.874)=2.950, p=.008). 

We also tested for a negative correlation between MorphScore and perplexity, such that better MorphScores were correlated with better performance. We fit a linear regression between the variables, but found no significant correlation (F(1, 13)=0.323, p=0.580). 

\subsection{Discussion}

One possible explanation for this result is that words in agglutinative languages are on average being segmented into more tokens, making it more likely that a token boundary will fall on a morpheme boundary. This in turn could be driven by word length, as agglutinative languages tend to have longer words. It could also be due to a higher number of token boundaries per word (fertility), as higher fertility means that as there are more token boundaries, it becomes more likely that one of the token boundaries would fall on a morpheme boundary due to chance. 
Upon analysis, we found that agglutinative languages indeed had longer words (t(29,923)=18.222, p$<$0.001) 
and more tokens per word (t(37,375)=34.27, p$<$0.001). 
We fit a linear regression with number of tokens per word, word length in characters, and morphological types as predictors for MorphScore. We found that fertility and word length are both negatively correlated with MorphScore ($\chi^2$(1)=61.457, p$<$0.001; $\chi^2$(1)=364.03, p$<$0.001; respectively); however, the effect sizes were extremely small with an adjusted $R^2=0.021$. Given these small effects, longer words or higher fertility cannot explain the greater than 20\% higher MorphScores for agglutinative languages. 

In order to mitigate concern about the choice to exclude one-token words from the calculation of MorphScore, we also calculate MorphScore such that a one-token word is counted as correct. Agglutinative languages still had higher MorphScores than fusional languages (t(18.874) = 2.393, p = 0.027). Furthermore, we found no difference in the absolute number of one-token words (t(19.867) = -0.768, p = 0.452) nor in the proportion of one-token words (t(17.014) = -0.577,  p =0.572)  between agglutinative and fusional languages.

These results are inconsistent with Hypothesis 1; morphological tokenizer alignment (as measured by MorphScore) is higher for agglutinative languages rather than lower, and this effect cannot be explained by higher fertility or longer word length. 

\section{H2: Tokenization Quality} \label{sec:rq2}

We next evaluate whether tokenization quality can explain the performance gap between agglutinative and fusional languages. We use two metrics of tokenization quality: compression and Rényi entropy. To achieve sufficient statistical power, we use the same tokenizers as the previous section, but add all the languages from \citet{chang2023multilinguality} with FLORES datasets and for which we have morphological type labels, for a total of 63 languages.
Perplexities for each tokenizer come from \citet{chang2023multilinguality}.

\subsection{Compression}

We use corpus token count (CTC; also known as sequence length) as our measure of compression. 
CTC \citep{schmidt2024tokenization} is the number of tokens it takes to encode a text. Lower CTC indicates better compression, which is thought to have various effects on performance, cost, and inference time. If a tokenizer encodes a given text with more tokens, this will mean more sequences in order to pass the text through a language model. Each sequence, thus, will contain less information. This leads to higher training cost and slower inference \citep{song-etal-2021-fast, petrov2024tokenpremium, yamaguchi2024empirical} and worse model performance \citep{galle-2019-investigating, liang-etal-2023-xlm, goldman2024unpacking}. 

We calculate CTC based on FLORES-200 \citep{nllb-22} by encoding the text for each language with its respective tokenizer and counting the sequence length, not including beginning- and end-of-sequence tokens. FLORES offers parallel texts for each language, meaning that each text contains the same content, and sequence lengths should be comparable between languages. 

\subsection{Rényi entropy}

Rényi entropy has been proposed as a metric of tokenization quality, as it measures the distribution of token frequencies over the tokenizer vocabulary, penalizing low- and high-frequency tokens. It has been shown to correlate with downstream performance \citep{zouhar-etal-2023-tokenization}. 

Rényi entropy might also capture undesirable tokenizer properties that could be causing the performance gap. Agglutinative languages have longer words \citep{fenk1999cognition, berg2022effects} and more unique word forms \citep{sandra1994morphology}. This means that a tokenizer with a fixed vocabulary size will necessarily use shorter tokens on average for an agglutinative language than for a fusional language\footnote{As there has not been previous empirical evidence to support this point, we test this. We use the same tokenizers as in the previous section and tokenize all the FLORES datasets for which we have corresponding monolingual tokenizers. We then calculate mean token length for the FLORES dataset. The mean token length for fusional languages was 2.92 characters and the mean token length for agglutinative languages was 3.25. This difference is statistically significant (t(68.36) = 3.236, p = 0.002).}. Shorter tokens will have higher frequencies on average \citep{berg2022effects}, and these tokens will carry less information, as the meaning of a word is distributed over more tokens. 

We calculate Rényi entropy from the FLORES dataset for each language using \texttt{tokenization-scorer}\footnote{\url{https://github.com/zouharvi/tokenization-scorer}} \citep{zouhar-etal-2023-tokenization} with the recommended setting ($\alpha$=2.5).

\subsection{Results}

Agglutinative languages have higher CTC (worse compression) than fusional languages (t(85.944)=2.507, p=0.014). On average, their sequences are 3.5\% longer. However, there is no correlation between CTC and perplexity (linear regression; F(1, 190)=2.05, p=0.154). This indicates that compression, at least measured in this way, does not explain the performance gap. 

There is also a difference in Rényi entropy between agglutinative and fusional languages. Agglutinative languages have worse (higher) Rényi entropy (M=0.547) than fusional languages (M=0.488; t(150.53)=5.168, p$<$0.001). 

In order to test whether Rényi entropy can help explain the performance gap, conduct a Likelihood Ratio Test comparing two linear mixed effects models. The full model predicts perplexity from morphological type, Rényi entropy, and CTC as fixed effects, with model size as a random intercept. We then fit a reduced model, removing morphological type as a fixed effect. We compare the models with an ANOVA and find that morphological type explains additional variance above and beyond the other predictors ($\chi^2$(3)=29.464, p$<$0.001). This indicates that Rényi entropy does not explain all of the variance significantly explained by morphological type. A variance partitioning analysis using the \texttt{partR2} package in R produces an $R^2$ for morphological type of 0.100 and for Rényi entropy of 0.030, while the full model $R^2$ is 0.144. Therefore the vast majority of the variance is still explained by morphological type. This suggests that Rényi entropy could explain only a small part of the performance gap. 

\subsection{Discussion}

These results are inconsistent with the hypothesis that tokenizer compression explains poorer language modeling performance for agglutinative languages. Other results, e.g. \citet{deletang2024language, schmidt2024tokenization}, also show a lack of relationship between compression and language model performance. While compression indicates how much information can be represented in a fixed sequence length, the effect of compression may be outweighed by other features of a particular tokenizer, or language models may be able to overcome suboptimal tokenization. This is an area for further research, as it remains unclear what the best criteria are for intrinsic evaluation of tokenizers \citep{zouhar-etal-2023-tokenization, chizhov2024picky}.

\section{H3: Data Measurement Disparities} \label{sec:rq3}

The final hypothesis for the performance gap is disparities in training data. 

The monolingual models used in \S\ref{sec:monolingual_evidence} and \S\ref{sec:rq2} were designed to be trained on comparable amounts of training data with comparable tokenizers \citep{chang2023multilinguality}. Nevertheless, there are differences in performance between languages. \citet{chang2024goldfish} trained a similar suite of models (the Goldfish models), taking into account the byte premiums for each language. 

Byte premiums \citep{arnett-etal-2024-bit} are the ratio of the number of bytes it takes to represent a content-matched text in different languages. For example, a text in a language with a byte premium of 3 relative to English will be three times larger in bytes than the content-matched English text file. One of the major contributors to byte premiums is the writing system used by a language. Latin characters are represented with a single byte in UTF-8 encoding. In the most extreme cases, characters for scripts like Khmer take three bytes per character, not including diacritics. As a result, some languages have byte premiums of up to 5 relative to English.

This has implications for many things, including how much text tokenizers are trained on. Most training data can be measured in number of tokens, but this isn't the case for tokenizer training data, as the tokenizer hasn't been trained yet. The Goldfish tokenizers and models are trained on byte-premium-scaled text quantities, which was designed to reduce the effects of the data measurement disparities between languages. 

In this section, we test whether taking byte premiums into account can reduce or completely eliminate the performance gap. We annotate 154 languages for morphological type and use the same procedure as in \S\ref{sec:perf_gap} to test for the performance gap with the Goldfish models. 

\subsection{Results}

The Goldfish models exhibit numerically higher perplexity for  agglutinative (M=143.62) than fusional languages (M=132.63), but this difference is not statistically significant (t(137.36)=1.180, p=0.077). Therefore, after taking byte premiums into effect, the Goldfish models do not exhibit the same performance gap that was demonstrated in previous research and in Section \ref{sec:perf_gap} above. 

We tested whether there was a relationship between byte premium and morphological type, and found that there was a marginally significant difference between byte premiums for agglutinative and fusional languages (\textit{t}-test; t(157.9)=1.960, p=0.0518).   
Based on these results, we argue that accounting for byte premiums by scaling training data reduces most of the variance previously accounted for by morphological type.

\section{Discussion}

We find that byte premiums explain the largest portion of the performance gap, which means that cross-lingual differences in text encoding size can explain these particular, previously documented performance differences. 
The results do not support the idea that some languages are harder to model than others, but it does seem that languages need to be treated differently, e.g. by scaling data quantities.
This result can be used to inform how much data should be used to train tokenizers and language models, especially in low-resource or multilingual settings. By not taking byte premiums into account, we may be disadvantaging languages which are historically under-represented in the field, even when resources for them do exist. 

While these results may be surprising based on the NLP literature, these results are consistent with evidence from language acquisition work, which has not shown any cross-linguistic differences in learnability of languages. These results are unsurprising from an empirical perspective, as work in Linguistics and NLP has consistently shown that more data will always facilitate better learning, irrespective of the complexity of the language. 

There do seem to be limits to the learnability of linguistic systems. There are some language systems that linguistic theory predicts are impossible for humans to learn. Recent work has shown that language models are less successful at learning those languages, compared to existing and possible linguistic systems \citep{kallini2024mission}. Therefore, we do not predict that these results hold for systems that are more complex than any attested natural language.

Recent work by \citet{kallini2024mission} has shown that language models struggle to learn languages that are more complex in ways that linguistic theory predicts to be impossible for humans to learn.

The results relating to Rényi entropy do suggest that there may be differences in tokenization which could be affecting performance, however more work is needed on this topic.

\section{Conclusion}

This paper first presented new evidence consistent with a performance gap between languages of different morphological types. We presented and tested three hypotheses as to the cause(s) for this performance gap: morphological alignment of the tokenizer, tokenization quality, and measurement disparities of dataset size. We found that while there was evidence that tokenization quality (as measured by Rényi entropy) plays a small role, dataset size seems to explain a large portion of the performance gap. After scaling training data according to byte premiums -- a measure of how many bytes it takes to represent text in different languages -- the performance gap goes away. 

To do this work, we also created MorphScore, which is an evaluation method that can be used to evaluate the morphological alignment of tokenizers. We release the datasets needed to evaluate MorphScore in 22 languages: \url{https://github.com/catherinearnett/morphscore}.  

These results raise questions about other unintended differences in the way languages are treated that could lead to differences in performance between languages. This is a critical issue for achieving language-general NLP systems and making language models perform equitably. While it does not seem that morphological typology is the primary reason for the observed performance gap, the initial observation led to greater understanding of crosslinguistic NLP. It is important to keep evaluating the dimensions along which languages vary and consider whether language technologies, such as LLMs, introduce inequalities between languages. We have yet to fully understand all the ways in which English-centric practices in NLP may have impeded progress for language models in other languages.   

\section*{Limitations}

For all of the analyses, we were limited by the number of languages for which we had morphological type annotations. These annotations are time-consuming and are themselves limited by the resources available, namely grammars and linguistic descriptions. The number of languages in the MorphScore analysis is even more limited. Having more annotations and datasets included in this work would make the analyses more reliable. This is an important place for expansion in future work.

In the MorphScore datasets, we were also limited by the type of existing data. There were differences in domain and breadth in the Universal Dependencies and UniMorph datasets for each language. There are also different numbers of items in each dataset for each language. This means some languages will have more diversity among the items and there will be more statistical power than others, therefore the treatment of each language was not the same, which could introduce uncontrolled variance. Additionally, the morpheme boundaries that are annotated for different languages was not consistent. Some boundaries were inflectional and some were derivational. If there existed large datasets of both inflectional and derivational morphologically annotated words in a wide range of languages, this would have improved the robustness of the MorphScore results. 

Finally, because the annotations in UD and UniMorph chose only one boundary (or, if there were multiple boundaries, we chose one), we can only evaluate whether the token boundaries align with the morpheme boundary we chose. We did this to limit confounds, as all but one datasets had one annotated boundary per word. Additionally, agglutinative languages would have more morpheme boundaries per word, which could skew results. However, there was no controlled selection process for which morpheme boundary was used for the MorphScore analysis, therefore this could have also affected results. 

In the analysis of the Goldfish models, the evidence that byte premiums account for the performance gap is supported by a marginally significant difference between byte premiums according to their morphological type. It is possible that with an even larger sample of languages, the effect would instead meet the standard threshold for significance. We argue that in conjunction with the other results, it still demonstrates that taking byte premiums into account significantly reduces the performance gap.

\section*{Acknowledgments}
We would like to thank the UC San Diego Social Sciences Computing Facility Team for the use of the Social Sciences Research and Development Environment (SSRDE) cluster. 
Experiments were conducted using hardware provided by the NVIDIA Corporation as part of an NVIDIA Academic Hardware Grant.

\bibliography{coling}

\begin{thebibliography}{107}
\providecommand{\natexlab}[1]{#1}

\bibitem[{Abulimiti and Schultz(2020)}]{abulimiti-schultz-2020-building}
Ayimunishagu Abulimiti and Tanja Schultz. 2020.
\newblock \href {https://aclanthology.org/2020.sltu-1.38} {Building language models for morphological rich low-resource languages using data from related donor languages: the case of {U}yghur}.
\newblock In \emph{Proceedings of the 1st Joint Workshop on Spoken Language Technologies for Under-resourced languages (SLTU) and Collaboration and Computing for Under-Resourced Languages (CCURL)}, pages 271--276, Marseille, France. European Language Resources association.

\bibitem[{Ackerman and Malouf(2013)}]{ackerman2013morphological}
Farrell Ackerman and Robert Malouf. 2013.
\newblock Morphological organization: The low conditional entropy conjecture.
\newblock \emph{Language}, 89(3):429--464.

\bibitem[{Adelani et~al.(2024)Adelani, Liu, Shen, Vassilyev, Alabi, Mao, Gao, and Lee}]{adelani-etal-2024-sib}
David Adelani, Hannah Liu, Xiaoyu Shen, Nikita Vassilyev, Jesujoba Alabi, Yanke Mao, Haonan Gao, and En-Shiun Lee. 2024.
\newblock \href {https://aclanthology.org/2024.eacl-long.14} {{SIB}-200: A simple, inclusive, and big evaluation dataset for topic classification in 200+ languages and dialects}.
\newblock In \emph{Proceedings of the 18th Conference of the European Chapter of the Association for Computational Linguistics (Volume 1: Long Papers)}, pages 226--245, St. Julian{'}s, Malta. Association for Computational Linguistics.

\bibitem[{{Aranes, Glyd Jun and Zeman, Dan }(2021)}]{ud_cebuano_gja}
{Aranes, Glyd Jun and Zeman, Dan }. 2021.
\newblock \href {https://github.com/UniversalDependencies/UD_Cebuano-GJA/tree/master} {Ud cebuano-gja}.
\newblock Accessed: 2024-09-04.

\bibitem[{Aranzabe et~al.(2015)Aranzabe, Atutxa, Bengoetxea, Gojenola, and Uria}]{aranzabe2015automatic}
Maria~Jesus Aranzabe, Aitziber Atutxa, Kepa Bengoetxea, Koldo Gojenola, and Larraitz Uria. 2015.
\newblock Automatic conversion of the basque dependency treebank to universal dependencies.
\newblock In \emph{Proceedings of the fourteenth international workshop on treebanks an linguistic theories (TLT14)}, pages 233--241.

\bibitem[{Arnett et~al.(2024{\natexlab{a}})Arnett, Chang, and Bergen}]{arnett-etal-2024-bit}
Catherine Arnett, Tyler~A. Chang, and Benjamin Bergen. 2024{\natexlab{a}}.
\newblock \href {https://aclanthology.org/2024.sigul-1.1} {A bit of a problem: Measurement disparities in dataset sizes across languages}.
\newblock In \emph{Proceedings of the 3rd Annual Meeting of the Special Interest Group on Under-resourced Languages @ LREC-COLING 2024}, pages 1--9, Torino, Italia. ELRA and ICCL.

\bibitem[{Arnett et~al.(2024{\natexlab{b}})Arnett, Rivière, Chang, and Trott}]{arnett-etal-2024-different}
Catherine Arnett, Pamela~D Rivière, Tyler Chang, and Sean Trott. 2024{\natexlab{b}}.
\newblock \href {https://doi.org/10.18653/v1/2024.sigmorphon-1.4} {Different tokenization schemes lead to comparable performance in {S}panish number agreement}.
\newblock In \emph{Proceedings of the 21st SIGMORPHON workshop on Computational Research in Phonetics, Phonology, and Morphology}, pages 32--38, Mexico City, Mexico. Association for Computational Linguistics.

\bibitem[{Atari et~al.(2023)Atari, Xue, Park, Blasi, and Henrich}]{atari2023humans}
Mohammad Atari, Mona~J Xue, Peter~S Park, Dami{\'a}n Blasi, and Joseph Henrich. 2023.
\newblock \href {https://osf.io/preprints/psyarxiv/5b26t?trk=public_post-text} {Which humans?}

\bibitem[{Baker(1996)}]{baker1996polysynthesis}
Mark Baker. 1996.
\newblock \emph{The polysynthesis parameter}.
\newblock Oxford Studies in Comparative Syntax. Oxford University Press.

\bibitem[{Banerjee and Bhattacharyya(2018)}]{banerjee-bhattacharyya-2018-meaningless}
Tamali Banerjee and Pushpak Bhattacharyya. 2018.
\newblock \href {https://doi.org/10.18653/v1/W18-1207} {Meaningless yet meaningful: Morphology grounded subword-level {NMT}}.
\newblock In \emph{Proceedings of the Second Workshop on Subword/Character {LE}vel Models}, pages 55--60, New Orleans. Association for Computational Linguistics.

\bibitem[{Bates(2010)}]{bates2010lme4}
Douglas~M Bates. 2010.
\newblock {lme4: Mixed-effects modeling with R}.

\bibitem[{Bauwens and Delobelle(2024)}]{bauwens-delobelle-2024-bpe}
Thomas Bauwens and Pieter Delobelle. 2024.
\newblock \href {https://doi.org/10.18653/v1/2024.naacl-long.324} {{BPE}-knockout: Pruning pre-existing {BPE} tokenisers with backwards-compatible morphological semi-supervision}.
\newblock In \emph{Proceedings of the 2024 Conference of the North American Chapter of the Association for Computational Linguistics: Human Language Technologies (Volume 1: Long Papers)}, pages 5810--5832, Mexico City, Mexico. Association for Computational Linguistics.

\bibitem[{Baxi and Bhatt(2021)}]{baxi2021morpheme}
Jatayu Baxi and Brijesh Bhatt. 2021.
\newblock \href {https://aclanthology.org/2021.icon-main.45} {Morpheme boundary detection {\&} grammatical feature prediction for {G}ujarati : Dataset {\&} model}.
\newblock In \emph{Proceedings of the 18th International Conference on Natural Language Processing (ICON)}, pages 369--377, National Institute of Technology Silchar, Silchar, India. NLP Association of India (NLPAI).

\bibitem[{Bender(2011)}]{bender2011achieving}
Emily~M Bender. 2011.
\newblock \href {https://citeseerx.ist.psu.edu/document?repid=rep1&type=pdf&doi=105ed573024e9a31eddc766b6018297ab4383bb9} {{On achieving and evaluating language-independence in NLP}}.
\newblock \emph{Linguistic Issues in Language Technology}, 6.

\bibitem[{Berg et~al.(2022)Berg, Z{\"o}rnig, and Lehr}]{berg2022effects}
Thomas Berg, Peter Z{\"o}rnig, and Charlotte Lehr. 2022.
\newblock \href {https://www.degruyter.com/document/doi/10.1515/glot-2022-2007/html} {The effects of type and token frequency on word length: a cross-linguistic study}.
\newblock \emph{Glottotheory}, 13(2):173--209.

\bibitem[{Bhat et~al.(2017)Bhat, Bhatt, Farudi, Klassen, Narasimhan, Palmer, Rambow, Sharma, Vaidya, Vishnu et~al.}]{bhathindi2017}
Riyaz~Ahmad Bhat, Rajesh Bhatt, Annahita Farudi, Prescott Klassen, Bhuvana Narasimhan, Martha Palmer, Owen Rambow, Dipti~Misra Sharma, Ashwini Vaidya, Sri~Ramagurumurthy Vishnu, et~al. 2017.
\newblock The hindi/urdu treebank project.
\newblock In \emph{Handbook of Linguistic Annotation}. Springer Press.

\bibitem[{Blasi et~al.(2022)Blasi, Anastasopoulos, and Neubig}]{blasi-etal-2022-systematic}
Damian Blasi, Antonios Anastasopoulos, and Graham Neubig. 2022.
\newblock \href {https://doi.org/10.18653/v1/2022.acl-long.376} {Systematic inequalities in language technology performance across the world{'}s languages}.
\newblock In \emph{Proceedings of the 60th Annual Meeting of the Association for Computational Linguistics (Volume 1: Long Papers)}, pages 5486--5505, Dublin, Ireland. Association for Computational Linguistics.

\bibitem[{Bostrom and Durrett(2020)}]{bostrom-durrett-2020-byte}
Kaj Bostrom and Greg Durrett. 2020.
\newblock \href {https://doi.org/10.18653/v1/2020.findings-emnlp.414} {Byte pair encoding is suboptimal for language model pretraining}.
\newblock In \emph{Findings of the Association for Computational Linguistics: EMNLP 2020}, pages 4617--4624, Online. Association for Computational Linguistics.

\bibitem[{Botev et~al.(2022)Botev, McCarthy, Wu, and Yarowsky}]{botev-etal-2022-deciphering}
Georgie Botev, Arya~D. McCarthy, Winston Wu, and David Yarowsky. 2022.
\newblock \href {https://aclanthology.org/2022.coling-1.472} {Deciphering and characterizing out-of-vocabulary words for morphologically rich languages}.
\newblock In \emph{Proceedings of the 29th International Conference on Computational Linguistics}, pages 5309--5326, Gyeongju, Republic of Korea. International Committee on Computational Linguistics.

\bibitem[{Brown(2010)}]{brown2010morphological}
Dunstan~Patrick Brown. 2010.
\newblock Morphological typology.
\newblock In \emph{Handbook of Linguistic Typology}, pages 487--503. Oxford University Press.

\bibitem[{Chang et~al.(2023)Chang, Arnett, Tu, and Bergen}]{chang2023multilinguality}
Tyler~A Chang, Catherine Arnett, Zhuowen Tu, and Benjamin~K Bergen. 2023.
\newblock \href {https://arxiv.org/abs/2311.09205} {{When Is Multilinguality a Curse? Language Modeling for 250 High-and Low-Resource Languages}}.
\newblock \emph{arXiv preprint arXiv:2311.09205}.

\bibitem[{Chang et~al.(2024)Chang, Arnett, Tu, and Bergen}]{chang2024goldfish}
Tyler~A Chang, Catherine Arnett, Zhuowen Tu, and Benjamin~K Bergen. 2024.
\newblock \href {https://arxiv.org/abs/2408.10441} {{Goldfish: Monolingual Language Models for 350 Languages}}.
\newblock \emph{arXiv preprint arXiv:2408.10441}.

\bibitem[{Chizhov et~al.(2024)Chizhov, Arnett, Korotkova, and Yamshchikov}]{chizhov2024picky}
Pavel Chizhov, Catherine Arnett, Elizaveta Korotkova, and Ivan~P. Yamshchikov. 2024.
\newblock \href {https://arxiv.org/abs/arXiv:2409.04599} {{BPE Gets Picky: Efficient Vocabulary Refinement During Tokenizer Training}}.
\newblock \emph{Preprint}, arXiv:arXiv:2409.04599.
\newblock Preprint.

\bibitem[{Choo and Kim(2023)}]{choo2023study}
Sanghyun Choo and Wonjoon Kim. 2023.
\newblock A study on the evaluation of tokenizer performance in natural language processing.
\newblock \emph{Applied Artificial Intelligence}, 37(1):2175112.

\bibitem[{Choudhury(2023)}]{choudhury2023generative}
Monojit Choudhury. 2023.
\newblock \href {https://www.nature.com/articles/s41562-023-01716-4} {{Generative AI has a language problem}}.
\newblock \emph{Nature Human Behaviour}, 7(11):1802--1803.

\bibitem[{Chun et~al.(2018)Chun, Han, Hwang, and Choi}]{chun-etal-2018-building}
Jayeol Chun, Na-Rae Han, Jena~D. Hwang, and Jinho~D. Choi. 2018.
\newblock \href {https://aclanthology.org/L18-1347} {Building {U}niversal {D}ependency treebanks in {K}orean}.
\newblock In \emph{Proceedings of the Eleventh International Conference on Language Resources and Evaluation ({LREC} 2018)}, Miyazaki, Japan. European Language Resources Association (ELRA).

\bibitem[{Conneau et~al.(2018)Conneau, Rinott, Lample, Williams, Bowman, Schwenk, and Stoyanov}]{conneau2018xnli}
Alexis Conneau, Ruty Rinott, Guillaume Lample, Adina Williams, Samuel Bowman, Holger Schwenk, and Veselin Stoyanov. 2018.
\newblock \href {https://doi.org/10.18653/v1/D18-1269} {{XNLI}: Evaluating cross-lingual sentence representations}.
\newblock In \emph{Proceedings of the 2018 Conference on Empirical Methods in Natural Language Processing}, pages 2475--2485, Brussels, Belgium. Association for Computational Linguistics.

\bibitem[{Cotterell et~al.(2018)Cotterell, Mielke, Eisner, and Roark}]{cotterell-etal-2018-languages}
Ryan Cotterell, Sabrina~J. Mielke, Jason Eisner, and Brian Roark. 2018.
\newblock \href {https://doi.org/10.18653/v1/N18-2085} {Are all languages equally hard to language-model?}
\newblock In \emph{Proceedings of the 2018 Conference of the North {A}merican Chapter of the Association for Computational Linguistics: Human Language Technologies, Volume 2 (Short Papers)}, pages 536--541, New Orleans, Louisiana. Association for Computational Linguistics.

\bibitem[{Dagan et~al.(2024)Dagan, Synnaeve, and Roziere}]{dagan2024getting}
Gautier Dagan, Gabriel Synnaeve, and Baptiste Roziere. 2024.
\newblock \href {https://openreview.net/forum?id=ZFYBnLljtT} {Getting the most out of your tokenizer for pre-training and domain adaptation}.
\newblock In \emph{Forty-first International Conference on Machine Learning}.

\bibitem[{Deletang et~al.(2024)Deletang, Ruoss, Duquenne, Catt, Genewein, Mattern, Grau-Moya, Wenliang, Aitchison, Orseau, Hutter, and Veness}]{deletang2024language}
Gregoire Deletang, Anian Ruoss, Paul-Ambroise Duquenne, Elliot Catt, Tim Genewein, Christopher Mattern, Jordi Grau-Moya, Li~Kevin Wenliang, Matthew Aitchison, Laurent Orseau, Marcus Hutter, and Joel Veness. 2024.
\newblock \href {https://openreview.net/forum?id=jznbgiynus} {Language modeling is compression}.
\newblock In \emph{The Twelfth International Conference on Learning Representations}.

\bibitem[{Dobrovoljc et~al.(2017)Dobrovoljc, Erjavec, and Krek}]{dobrovoljc-etal-2017-universal}
Kaja Dobrovoljc, Toma{\v{z}} Erjavec, and Simon Krek. 2017.
\newblock \href {https://doi.org/10.18653/v1/W17-1406} {The {U}niversal {D}ependencies treebank for {S}lovenian}.
\newblock In \emph{Proceedings of the 6th Workshop on {B}alto-{S}lavic Natural Language Processing}, pages 33--38, Valencia, Spain. Association for Computational Linguistics.

\bibitem[{Dobrovoljc and Ljube{\v{s}}i{\'c}(2022)}]{dobrovoljc-ljubesic-2022-extending}
Kaja Dobrovoljc and Nikola Ljube{\v{s}}i{\'c}. 2022.
\newblock \href {https://aclanthology.org/2022.law-1.3} {Extending the {SSJ} {U}niversal {D}ependencies treebank for {S}lovenian: Was it worth it?}
\newblock In \emph{Proceedings of the 16th Linguistic Annotation Workshop (LAW-XVI) within LREC2022}, pages 15--22, Marseille, France. European Language Resources Association.

\bibitem[{Dressler(2010)}]{dressler2010typological}
Wolfgang~U Dressler. 2010.
\newblock A typological approach to first language acquisition.
\newblock \emph{Language acquisition across linguistic and cognitive systems}, 52:109--124.

\bibitem[{Dryer and Haspelmath(2013)}]{dryer2013wals}
Matthew~S. Dryer and Martin Haspelmath. 2013.
\newblock \href {https://doi.org/10.5281/zenodo.7385533} {{WALS Online (v2020.3)}}.

\bibitem[{Fenk-Oczlon and Fenk(1999)}]{fenk1999cognition}
Gertraud Fenk-Oczlon and August Fenk. 1999.
\newblock Cognition, quantitative linguistics, and systemic typology.
\newblock \emph{Linguistic Typology}, 3:151--177.

\bibitem[{Gall{\'e}(2019)}]{galle-2019-investigating}
Matthias Gall{\'e}. 2019.
\newblock \href {https://doi.org/10.18653/v1/D19-1141} {Investigating the effectiveness of {BPE}: The power of shorter sequences}.
\newblock In \emph{Proceedings of the 2019 Conference on Empirical Methods in Natural Language Processing and the 9th International Joint Conference on Natural Language Processing (EMNLP-IJCNLP)}, pages 1375--1381, Hong Kong, China. Association for Computational Linguistics.

\bibitem[{Gerz et~al.(2018{\natexlab{a}})Gerz, Vuli{\'c}, Ponti, Naradowsky, Reichart, and Korhonen}]{gerz2018language}
Daniela Gerz, Ivan Vuli{\'c}, Edoardo Ponti, Jason Naradowsky, Roi Reichart, and Anna Korhonen. 2018{\natexlab{a}}.
\newblock \href {https://doi.org/10.1162/tacl_a_00032} {Language modeling for morphologically rich languages: Character-aware modeling for word-level prediction}.
\newblock \emph{Transactions of the Association for Computational Linguistics}, 6:451--465.

\bibitem[{Gerz et~al.(2018{\natexlab{b}})Gerz, Vuli{\'c}, Ponti, Reichart, and Korhonen}]{gerz-etal-2018-relation}
Daniela Gerz, Ivan Vuli{\'c}, Edoardo~Maria Ponti, Roi Reichart, and Anna Korhonen. 2018{\natexlab{b}}.
\newblock \href {https://doi.org/10.18653/v1/D18-1029} {On the relation between linguistic typology and (limitations of) multilingual language modeling}.
\newblock In \emph{Proceedings of the 2018 Conference on Empirical Methods in Natural Language Processing}, pages 316--327, Brussels, Belgium. Association for Computational Linguistics.

\bibitem[{Goldman et~al.(2024)Goldman, Caciularu, Eyal, Cao, Szpektor, and Tsarfaty}]{goldman2024unpacking}
Omer Goldman, Avi Caciularu, Matan Eyal, Kris Cao, Idan Szpektor, and Reut Tsarfaty. 2024.
\newblock \href {https://arxiv.org/pdf/2403.06265} {Unpacking {T}okenization: {E}valuating {T}ext {C}ompression and its {C}orrelation with {M}odel {P}erformance}.
\newblock \emph{arXiv preprint arXiv:2403.06265}.

\bibitem[{Gueta et~al.(2023)Gueta, Goldman, and Tsarfaty}]{gueta2023explicit}
Eylon Gueta, Omer Goldman, and Reut Tsarfaty. 2023.
\newblock {Explicit Morphological Knowledge Improves Pre-training of Language Models for Hebrew}.
\newblock \emph{arXiv e-prints}, pages arXiv--2311.

\bibitem[{Guo et~al.(2020)Guo, Dai, Vrande{\v{c}}i{\'c}, and Al-Rfou}]{guo-etal-2020-wiki}
Mandy Guo, Zihang Dai, Denny Vrande{\v{c}}i{\'c}, and Rami Al-Rfou. 2020.
\newblock \href {https://aclanthology.org/2020.lrec-1.297} {{W}iki-40{B}: Multilingual language model dataset}.
\newblock In \emph{Proceedings of the Twelfth Language Resources and Evaluation Conference}, pages 2440--2452, Marseille, France. European Language Resources Association.

\bibitem[{Gutierrez-Vasques et~al.(2023)Gutierrez-Vasques, Bentz, and Samard{\v{z}}i{\'c}}]{gutierrez2023languages}
Ximena Gutierrez-Vasques, Christian Bentz, and Tanja Samard{\v{z}}i{\'c}. 2023.
\newblock \href {https://doi.org/10.1162/coli_a_00489} {Languages through the looking glass of {BPE} compression}.
\newblock \emph{Computational Linguistics}, 49(4):943--1001.

\bibitem[{Hakkani-T{\"u}r et~al.(2002)Hakkani-T{\"u}r, Oflazer, and T{\"u}r}]{hakkani2002statistical}
Dilek~Z Hakkani-T{\"u}r, Kemal Oflazer, and G{\"o}khan T{\"u}r. 2002.
\newblock Statistical morphological disambiguation for agglutinative languages.
\newblock \emph{Computers and the Humanities}, 36:381--410.

\bibitem[{Haspelmath(2009)}]{haspelmath2009empirical}
Martin Haspelmath. 2009.
\newblock \href {https://link.springer.com/chapter/10.1007/978-1-4020-8825-4_2} {{An empirical test of the Agglutination Hypothesis}}.
\newblock \emph{Universals of language today}, pages 13--29.

\bibitem[{Hoffmann et~al.(2024)Hoffmann, Borgeaud, Mensch, Buchatskaya, Cai, Rutherford, de~Las~Casas, Hendricks, Welbl, Clark, Hennigan, Noland, Millican, van~den Driessche, Damoc, Guy, Osindero, Simonyan, Elsen, Vinyals, Rae, and Sifre}]{hoffmann2022training}
Jordan Hoffmann, Sebastian Borgeaud, Arthur Mensch, Elena Buchatskaya, Trevor Cai, Eliza Rutherford, Diego de~Las~Casas, Lisa~Anne Hendricks, Johannes Welbl, Aidan Clark, Tom Hennigan, Eric Noland, Katie Millican, George van~den Driessche, Bogdan Damoc, Aurelia Guy, Simon Osindero, Karen Simonyan, Erich Elsen, Oriol Vinyals, Jack~W. Rae, and Laurent Sifre. 2024.
\newblock \href {https://dl.acm.org/doi/abs/10.5555/3600270.3602446} {Training compute-optimal large language models}.
\newblock In \emph{Proceedings of the 36th International Conference on Neural Information Processing Systems}, NIPS '22, Red Hook, NY, USA. Curran Associates Inc.

\bibitem[{Hofmann et~al.(2021)Hofmann, Pierrehumbert, and Sch{\"u}tze}]{hofmann-etal-2021-superbizarre}
Valentin Hofmann, Janet Pierrehumbert, and Hinrich Sch{\"u}tze. 2021.
\newblock \href {https://doi.org/10.18653/v1/2021.acl-long.279} {Superbizarre is not superb: Derivational morphology improves {BERT}{'}s interpretation of complex words}.
\newblock In \emph{Proceedings of the 59th Annual Meeting of the Association for Computational Linguistics and the 11th International Joint Conference on Natural Language Processing (Volume 1: Long Papers)}, pages 3594--3608, Online. Association for Computational Linguistics.

\bibitem[{Hofmann et~al.(2022)Hofmann, Schuetze, and Pierrehumbert}]{hofmann-etal-2022-embarrassingly}
Valentin Hofmann, Hinrich Schuetze, and Janet Pierrehumbert. 2022.
\newblock \href {https://doi.org/10.18653/v1/2022.acl-short.43} {An embarrassingly simple method to mitigate undesirable properties of pretrained language model tokenizers}.
\newblock In \emph{Proceedings of the 60th Annual Meeting of the Association for Computational Linguistics (Volume 2: Short Papers)}, pages 385--393, Dublin, Ireland. Association for Computational Linguistics.

\bibitem[{Jabbar(2023)}]{jabbar2023morphpiece}
Haris Jabbar. 2023.
\newblock \href {https://arxiv.org/abs/2307.07262} {{MorphPiece: Moving away from Statistical Language Representation}}.
\newblock \emph{arXiv preprint arXiv:2307.07262}.

\bibitem[{Joshi et~al.(2020)Joshi, Santy, Budhiraja, Bali, and Choudhury}]{joshi-etal-2020-state}
Pratik Joshi, Sebastin Santy, Amar Budhiraja, Kalika Bali, and Monojit Choudhury. 2020.
\newblock \href {https://doi.org/10.18653/v1/2020.acl-main.560} {The state and fate of linguistic diversity and inclusion in the {NLP} world}.
\newblock In \emph{Proceedings of the 58th Annual Meeting of the Association for Computational Linguistics}, pages 6282--6293, Online. Association for Computational Linguistics.

\bibitem[{Kallini et~al.(2024)Kallini, Papadimitriou, Futrell, Mahowald, and Potts}]{kallini2024mission}
Julie Kallini, Isabel Papadimitriou, Richard Futrell, Kyle Mahowald, and Christopher Potts. 2024.
\newblock Mission: Impossible language models.
\newblock \emph{arXiv preprint arXiv:2401.06416}.

\bibitem[{Kaya and Tantu{\u{g}}(2024)}]{kaya2024effect}
Yi{\u{g}}it~Bekir Kaya and A~C{\"u}neyd Tantu{\u{g}}. 2024.
\newblock \href {https://www.sciencedirect.com/science/article/pii/S2667305324000115} {{Effect of tokenization granularity for Turkish large language models}}.
\newblock \emph{Intelligent Systems with Applications}, 21:200335.

\bibitem[{Khanuja et~al.(2023)Khanuja, Ruder, and Talukdar}]{khanuja-etal-2023-evaluating}
Simran Khanuja, Sebastian Ruder, and Partha Talukdar. 2023.
\newblock \href {https://doi.org/10.18653/v1/2023.findings-eacl.131} {Evaluating the diversity, equity, and inclusion of {NLP} technology: A case study for {I}ndian languages}.
\newblock In \emph{Findings of the Association for Computational Linguistics: EACL 2023}, pages 1763--1777, Dubrovnik, Croatia. Association for Computational Linguistics.

\bibitem[{Kirov et~al.(2018)Kirov, Cotterell, Sylak-Glassman, Walther, Vylomova, Xia, Faruqui, Mielke, McCarthy, K{\"u}bler, Yarowsky, Eisner, and Hulden}]{kirov-etal-2018-unimorph}
Christo Kirov, Ryan Cotterell, John Sylak-Glassman, G{\'e}raldine Walther, Ekaterina Vylomova, Patrick Xia, Manaal Faruqui, Sabrina~J. Mielke, Arya McCarthy, Sandra K{\"u}bler, David Yarowsky, Jason Eisner, and Mans Hulden. 2018.
\newblock \href {https://aclanthology.org/L18-1293} {{U}ni{M}orph 2.0: {U}niversal {M}orphology}.
\newblock In \emph{Proceedings of the Eleventh International Conference on Language Resources and Evaluation ({LREC} 2018)}, Miyazaki, Japan. European Language Resources Association (ELRA).

\bibitem[{Klein and Tsarfaty(2020)}]{klein-tsarfaty-2020-getting}
Stav Klein and Reut Tsarfaty. 2020.
\newblock \href {https://doi.org/10.18653/v1/2020.sigmorphon-1.24} {Getting the {\#}{\#}life out of living: How adequate are word-pieces for modelling complex morphology?}
\newblock In \emph{Proceedings of the 17th SIGMORPHON Workshop on Computational Research in Phonetics, Phonology, and Morphology}, pages 204--209, Online. Association for Computational Linguistics.

\bibitem[{Kudo and Richardson(2018)}]{kudo-richardson-2018-sentencepiece}
Taku Kudo and John Richardson. 2018.
\newblock \href {https://doi.org/10.18653/v1/D18-2012} {{S}entence{P}iece: A simple and language independent subword tokenizer and detokenizer for neural text processing}.
\newblock In \emph{Proceedings of the 2018 Conference on Empirical Methods in Natural Language Processing: System Demonstrations}, pages 66--71, Brussels, Belgium. Association for Computational Linguistics.

\bibitem[{Larasati et~al.(2011)Larasati, Kubo{\v{n}}, and Zeman}]{larasati2011indonesian}
Septina~Dian Larasati, Vladislav Kubo{\v{n}}, and Daniel Zeman. 2011.
\newblock Indonesian morphology tool (morphind): Towards an indonesian corpus.
\newblock In \emph{Systems and Frameworks for Computational Morphology: Second International Workshop, SFCM 2011, Zurich, Switzerland, August 26, 2011. Proceedings 2}, pages 119--129. Springer.

\bibitem[{Le~Scao et~al.(2023)Le~Scao, Fan, Akiki, Pavlick, Ili{\'c}, Hesslow, Castagn{\'e}, Luccioni, Yvon, Gall{\'e} et~al.}]{le2023bloom}
Teven Le~Scao, Angela Fan, Christopher Akiki, Ellie Pavlick, Suzana Ili{\'c}, Daniel Hesslow, Roman Castagn{\'e}, Alexandra~Sasha Luccioni, Fran{\c{c}}ois Yvon, Matthias Gall{\'e}, et~al. 2023.
\newblock Bloom: A 176b-parameter open-access multilingual language model.

\bibitem[{Lee et~al.(2024)Lee, Moon, Lee, Park, Eo, Ko, Seo, Lee, and Lim}]{lee-etal-2024-length}
Jungseob Lee, Hyeonseok Moon, Seungjun Lee, Chanjun Park, Sugyeong Eo, Hyunwoong Ko, Jaehyung Seo, Seungyoon Lee, and Heuiseok Lim. 2024.
\newblock \href {https://aclanthology.org/2024.findings-acl.135} {Length-aware byte pair encoding for mitigating over-segmentation in {K}orean machine translation}.
\newblock In \emph{Findings of the Association for Computational Linguistics ACL 2024}, pages 2287--2303, Bangkok, Thailand and virtual meeting. {Association for Computational Linguistics}.

\bibitem[{Liang et~al.(2023)Liang, Gonen, Mao, Hou, Goyal, Ghazvininejad, Zettlemoyer, and Khabsa}]{liang-etal-2023-xlm}
Davis Liang, Hila Gonen, Yuning Mao, Rui Hou, Naman Goyal, Marjan Ghazvininejad, Luke Zettlemoyer, and Madian Khabsa. 2023.
\newblock \href {https://doi.org/10.18653/v1/2023.emnlp-main.813} {{XLM}-{V}: {O}vercoming the {V}ocabulary {B}ottleneck in {M}ultilingual {M}asked {L}anguage {M}odels}.
\newblock In \emph{Proceedings of the 2023 Conference on Empirical Methods in Natural Language Processing}, pages 13142--13152, Singapore. Association for Computational Linguistics.

\bibitem[{Libovick{\`y} and Helcl(2024)}]{libovicky2024lexically}
Jind{\v{r}}ich Libovick{\`y} and Jind{\v{r}}ich Helcl. 2024.
\newblock Lexically grounded subword segmentation.
\newblock \emph{arXiv preprint arXiv:2406.13560}.

\bibitem[{Lin et~al.(2024)Lin, Ji, Tiedemann, Martins, and Sch{\"u}tze}]{lin2024mala}
Peiqin Lin, Shaoxiong Ji, J{\"o}rg Tiedemann, Andr{\'e}~FT Martins, and Hinrich Sch{\"u}tze. 2024.
\newblock \href {https://arxiv.org/abs/2401.13303} {{MaLA-500: Massive Language Adaptation of Large Language Models}}.
\newblock \emph{arXiv preprint arXiv:2401.13303}.

\bibitem[{Lin et~al.(2022)Lin, Mihaylov, Artetxe, Wang, Chen, Simig, Ott, Goyal, Bhosale, Du, Pasunuru, Shleifer, Koura, Chaudhary, O{'}Horo, Wang, Zettlemoyer, Kozareva, Diab, Stoyanov, and Li}]{lin-etal-2022-shot}
Xi~Victoria Lin, Todor Mihaylov, Mikel Artetxe, Tianlu Wang, Shuohui Chen, Daniel Simig, Myle Ott, Naman Goyal, Shruti Bhosale, Jingfei Du, Ramakanth Pasunuru, Sam Shleifer, Punit~Singh Koura, Vishrav Chaudhary, Brian O{'}Horo, Jeff Wang, Luke Zettlemoyer, Zornitsa Kozareva, Mona Diab, Veselin Stoyanov, and Xian Li. 2022.
\newblock \href {https://doi.org/10.18653/v1/2022.emnlp-main.616} {Few-shot learning with multilingual generative language models}.
\newblock In \emph{Proceedings of the 2022 Conference on Empirical Methods in Natural Language Processing}, pages 9019--9052, Abu Dhabi, United Arab Emirates. Association for Computational Linguistics.

\bibitem[{Lobzhanidze(2022)}]{lobzhanidze2022finite}
Irina Lobzhanidze. 2022.
\newblock \emph{Finite-State Computational Morphology}.
\newblock Springer.

\bibitem[{Lynn and Foster(2016)}]{lynn2016universal}
Teresa Lynn and Jennifer Foster. 2016.
\newblock {Universal Dependencies for Irish}.
\newblock In \emph{Proceedings of the Second Celtic Language Technology Workshop}, Paris, France.

\bibitem[{Mach{\'a}{\v{c}}ek et~al.(2018)Mach{\'a}{\v{c}}ek, Vidra, and Bojar}]{machavcek2018morphological}
Dominik Mach{\'a}{\v{c}}ek, Jon{\'a}{\v{s}} Vidra, and Ond{\v{r}}ej Bojar. 2018.
\newblock {Morphological and language-agnostic word segmentation for NMT}.
\newblock In \emph{International Conference on Text, Speech, and Dialogue}, pages 277--284. Springer.

\bibitem[{Matsuzaki et~al.(2024)Matsuzaki, Taniguchi, Inui, and Sakaguchi}]{matsuzaki-etal-2024-j}
Kosuke Matsuzaki, Masaya Taniguchi, Kentaro Inui, and Keisuke Sakaguchi. 2024.
\newblock \href {https://doi.org/10.18653/v1/2024.sigmorphon-1.2} {{J}-{U}ni{M}orph: {J}apanese morphological annotation through the universal feature schema}.
\newblock In \emph{Proceedings of the 21st SIGMORPHON workshop on Computational Research in Phonetics, Phonology, and Morphology}, pages 7--19, Mexico City, Mexico. Association for Computational Linguistics.

\bibitem[{Minixhofer et~al.(2023)Minixhofer, Pfeiffer, and Vuli{\'c}}]{minixhofer-etal-2023-compoundpiece}
Benjamin Minixhofer, Jonas Pfeiffer, and Ivan Vuli{\'c}. 2023.
\newblock \href {https://doi.org/10.18653/v1/2023.emnlp-main.24} {{C}ompound{P}iece: Evaluating and improving decompounding performance of language models}.
\newblock In \emph{Proceedings of the 2023 Conference on Empirical Methods in Natural Language Processing}, pages 343--359, Singapore. Association for Computational Linguistics.

\bibitem[{Muennighoff et~al.(2023)Muennighoff, Wang, Sutawika, Roberts, Biderman, Le~Scao, Bari, Shen, Yong, Schoelkopf, Tang, Radev, Aji, Almubarak, Albanie, Alyafeai, Webson, Raff, and Raffel}]{muennighoff-etal-2023-crosslingual}
Niklas Muennighoff, Thomas Wang, Lintang Sutawika, Adam Roberts, Stella Biderman, Teven Le~Scao, M~Saiful Bari, Sheng Shen, Zheng~Xin Yong, Hailey Schoelkopf, Xiangru Tang, Dragomir Radev, Alham~Fikri Aji, Khalid Almubarak, Samuel Albanie, Zaid Alyafeai, Albert Webson, Edward Raff, and Colin Raffel. 2023.
\newblock \href {https://doi.org/10.18653/v1/2023.acl-long.891} {{Crosslingual Generalization through Multitask Finetuning}}.
\newblock In \emph{Proceedings of the 61st Annual Meeting of the Association for Computational Linguistics (Volume 1: Long Papers)}, pages 15991--16111, Toronto, Canada. Association for Computational Linguistics.

\bibitem[{{NLLB Team} et~al.(2022){NLLB Team}, Costa-jussà, Cross, Çelebi, Elbayad, Heafield, Heffernan, Kalbassi, Lam, Licht, Maillard, Sun, Wang, Wenzek, Youngblood, Akula, Barrault, Mejia-Gonzalez, Hansanti, Hoffman, Jarrett, Sadagopan, Rowe, Spruit, Tran, Andrews, Ayan, Bhosale, Edunov, Fan, Gao, Goswami, Guzmán, Koehn, Mourachko, Ropers, Saleem, Schwenk, and Wang}]{nllb-22}
{NLLB Team}, Marta~R. Costa-jussà, James Cross, Onur Çelebi, Maha Elbayad, Kenneth Heafield, Kevin Heffernan, Elahe Kalbassi, Janice Lam, Daniel Licht, Jean Maillard, Anna Sun, Skyler Wang, Guillaume Wenzek, Al~Youngblood, Bapi Akula, Loic Barrault, Gabriel Mejia-Gonzalez, Prangthip Hansanti, John Hoffman, Semarley Jarrett, Kaushik~Ram Sadagopan, Dirk Rowe, Shannon Spruit, Chau Tran, Pierre Andrews, Necip~Fazil Ayan, Shruti Bhosale, Sergey Edunov, Angela Fan, Cynthia Gao, Vedanuj Goswami, Francisco Guzmán, Philipp Koehn, Alexandre Mourachko, Christophe Ropers, Safiyyah Saleem, Holger Schwenk, and Jeff Wang. 2022.
\newblock \href {https://arxiv.org/abs/arXiv:1902.01382} {No language left behind: Scaling human-centered machine translation}.

\bibitem[{Nzeyimana and Niyongabo~Rubungo(2022)}]{nzeyimana-niyongabo-rubungo-2022-kinyabert}
Antoine Nzeyimana and Andre Niyongabo~Rubungo. 2022.
\newblock \href {https://doi.org/10.18653/v1/2022.acl-long.367} {{K}inya{BERT}: a morphology-aware {K}inyarwanda language model}.
\newblock In \emph{Proceedings of the 60th Annual Meeting of the Association for Computational Linguistics (Volume 1: Long Papers)}, pages 5347--5363, Dublin, Ireland. Association for Computational Linguistics.

\bibitem[{Palmer et~al.(2009)Palmer, Bhatt, Narasimhan, Rambow, Sharma, and Xia}]{palmer2009hindi}
Martha Palmer, Rajesh Bhatt, Bhuvana Narasimhan, Owen Rambow, Dipti~Misra Sharma, and Fei Xia. 2009.
\newblock {Hindi syntax: Annotating dependency, lexical predicate-argument structure, and phrase structure}.
\newblock In \emph{The 7th International Conference on Natural Language Processing}, pages 14--17.

\bibitem[{Park et~al.(2021)Park, Zhang, Haley, Steimel, Liu, and Schwartz}]{park2021morphology}
Hyunji~Hayley Park, Katherine~J Zhang, Coleman Haley, Kenneth Steimel, Han Liu, and Lane Schwartz. 2021.
\newblock Morphology matters: A multilingual language modeling analysis.
\newblock \emph{Transactions of the Association for Computational Linguistics}, 9:261--276.

\bibitem[{Petrov et~al.(2023)Petrov, La~Malfa, Torr, and Bibi}]{petrov2024tokenpremium}
Aleksandar Petrov, Emanuele La~Malfa, Philip Torr, and Adel Bibi. 2023.
\newblock \href {https://proceedings.neurips.cc/paper_files/paper/2023/file/74bb24dca8334adce292883b4b651eda-Paper-Conference.pdf} {Language model tokenizers introduce unfairness between languages}.
\newblock In \emph{Advances in Neural Information Processing Systems}, volume~36, pages 36963--36990. Curran Associates, Inc.

\bibitem[{Pimentel et~al.(2021)Pimentel, Ryskina, Mielke, Wu, Chodroff, Leonard, Nicolai, Ghanggo~Ate, Khalifa, Habash, El-Khaissi, Goldman, Gasser, Lane, Coler, Oncevay, Montoya~Samame, Silva~Villegas, Ek, Bernardy, Shcherbakov, Bayyr-ool, Sheifer, Ganieva, Plugaryov, Klyachko, Salehi, Krizhanovsky, Krizhanovsky, Vania, Ivanova, Salchak, Straughn, Liu, Washington, Ataman, Kiera{\'s}, Woli{\'n}ski, Suhardijanto, Stoehr, Nuriah, Ratan, Tyers, Ponti, Aiton, Hatcher, Prud'hommeaux, Kumar, Hulden, Barta, Lakatos, Szolnok, {\'A}cs, Raj, Yarowsky, Cotterell, Ambridge, and Vylomova}]{pimentel-ryskina-etal-2021-sigmorphon}
Tiago Pimentel, Maria Ryskina, Sabrina~J. Mielke, Shijie Wu, Eleanor Chodroff, Brian Leonard, Garrett Nicolai, Yustinus Ghanggo~Ate, Salam Khalifa, Nizar Habash, Charbel El-Khaissi, Omer Goldman, Michael Gasser, William Lane, Matt Coler, Arturo Oncevay, Jaime~Rafael Montoya~Samame, Gema~Celeste Silva~Villegas, Adam Ek, Jean-Philippe Bernardy, Andrey Shcherbakov, Aziyana Bayyr-ool, Karina Sheifer, Sofya Ganieva, Matvey Plugaryov, Elena Klyachko, Ali Salehi, Andrew Krizhanovsky, Natalia Krizhanovsky, Clara Vania, Sardana Ivanova, Aelita Salchak, Christopher Straughn, Zoey Liu, Jonathan~North Washington, Duygu Ataman, Witold Kiera{\'s}, Marcin Woli{\'n}ski, Totok Suhardijanto, Niklas Stoehr, Zahroh Nuriah, Shyam Ratan, Francis~M. Tyers, Edoardo~M. Ponti, Grant Aiton, Richard~J. Hatcher, Emily Prud'hommeaux, Ritesh Kumar, Mans Hulden, Botond Barta, Dorina Lakatos, G{\'a}bor Szolnok, Judit {\'A}cs, Mohit Raj, David Yarowsky, Ryan Cotterell, Ben Ambridge, and Ekaterina Vylomova. 2021.
\newblock \href {https://doi.org/10.18653/v1/2021.sigmorphon-1.25} {{SIGMORPHON 2021 Shared Task on Morphological Reinflection: Generalization Across Languages}}.
\newblock In \emph{Proceedings of the 18th SIGMORPHON Workshop on Computational Research in Phonetics, Phonology, and Morphology}, pages 229--259, Online. Association for Computational Linguistics.

\bibitem[{Pinnis et~al.(2017)Pinnis, Kri{\v{s}}lauks, Deksne, and Miks}]{pinnis2017neural}
M{\=a}rcis Pinnis, Rihards Kri{\v{s}}lauks, Daiga Deksne, and Toms Miks. 2017.
\newblock Neural machine translation for morphologically rich languages with improved sub-word units and synthetic data.
\newblock In \emph{Text, Speech, and Dialogue: 20th International Conference, TSD 2017, Prague, Czech Republic, August 27-31, 2017, Proceedings 20}, pages 237--245. Springer.

\bibitem[{Plank(1991)}]{plank1991abundance}
Frans Plank. 1991.
\newblock {Of abundance and scantiness in inflection: A typological prelude}.
\newblock \emph{Paradigms: the economy of inflection}, pages 1--39.

\bibitem[{Plank(1999)}]{plank1999split}
Frans Plank. 1999.
\newblock Split morphology: How agglutination and flexion mix.
\newblock \emph{Linguistic Typology}, 3:279--340.

\bibitem[{Ponti et~al.(2020)Ponti, Glava{\v{s}}, Majewska, Liu, Vuli{\'c}, and Korhonen}]{ponti-etal-2020-xcopa}
Edoardo~Maria Ponti, Goran Glava{\v{s}}, Olga Majewska, Qianchu Liu, Ivan Vuli{\'c}, and Anna Korhonen. 2020.
\newblock \href {https://doi.org/10.18653/v1/2020.emnlp-main.185} {{XCOPA}: A multilingual dataset for causal commonsense reasoning}.
\newblock In \emph{Proceedings of the 2020 Conference on Empirical Methods in Natural Language Processing (EMNLP)}, pages 2362--2376, Online. Association for Computational Linguistics.

\bibitem[{Prokopidis and Papageorgiou(2014)}]{prokopidis-papageorgiou-2014-experiments}
Prokopis Prokopidis and Haris Papageorgiou. 2014.
\newblock \href {https://aclanthology.org/W14-6109} {Experiments for dependency parsing of {G}reek}.
\newblock In \emph{Proceedings of the First Joint Workshop on Statistical Parsing of Morphologically Rich Languages and Syntactic Analysis of Non-Canonical Languages}, pages 90--96, Dublin, Ireland. Dublin City University.

\bibitem[{Ramasamy and \v{Z}abokrtsk\'{y}(2012)}]{tamil_ud}
Loganathan Ramasamy and Zden\v{e}k \v{Z}abokrtsk\'{y}. 2012.
\newblock \href {http://www.lrec-conf.org/proceedings/lrec2012/summaries/456.html} {Prague dependency style treebank for {Tamil}}.
\newblock In \emph{Proceedings of Eighth International Conference on Language Resources and Evaluation ({LREC} 2012)}, pages 1888--1894, \.{I}stanbul, Turkey.

\bibitem[{Ramesh et~al.(2023)Ramesh, Sitaram, and Choudhury}]{ramesh-etal-2023-fairness}
Krithika Ramesh, Sunayana Sitaram, and Monojit Choudhury. 2023.
\newblock \href {https://doi.org/10.18653/v1/2023.findings-eacl.157} {Fairness in language models beyond {E}nglish: Gaps and challenges}.
\newblock In \emph{Findings of the Association for Computational Linguistics: EACL 2023}, pages 2106--2119, Dubrovnik, Croatia. Association for Computational Linguistics.

\bibitem[{Ranathunga and de~Silva(2022)}]{ranathunga-de-silva-2022-languages}
Surangika Ranathunga and Nisansa de~Silva. 2022.
\newblock \href {https://aclanthology.org/2022.aacl-main.62} {Some languages are more equal than others: Probing deeper into the linguistic disparity in the {NLP} world}.
\newblock In \emph{Proceedings of the 2nd Conference of the Asia-Pacific Chapter of the Association for Computational Linguistics and the 12th International Joint Conference on Natural Language Processing (Volume 1: Long Papers)}, pages 823--848, Online only. Association for Computational Linguistics.

\bibitem[{Rasooli et~al.(2013)Rasooli, Kouhestani, and Moloodi}]{rasooli-etal-2013-development}
Mohammad~Sadegh Rasooli, Manouchehr Kouhestani, and Amirsaeid Moloodi. 2013.
\newblock \href {https://aclanthology.org/N13-1031} {Development of a {P}ersian syntactic dependency treebank}.
\newblock In \emph{Proceedings of the 2013 Conference of the North {A}merican Chapter of the Association for Computational Linguistics: Human Language Technologies}, pages 306--314, Atlanta, Georgia. Association for Computational Linguistics.

\bibitem[{R{\"o}gnvaldsson et~al.(2012)R{\"o}gnvaldsson, Ingason, Sigur\ipa{\dh}sson, and Wallenberg}]{rognvaldsson-etal-2012-icelandic}
Eir{\'\i}kur R{\"o}gnvaldsson, Anton~Karl Ingason, Einar~Freyr Sigur\ipa{\dh}sson, and Joel Wallenberg. 2012.
\newblock \href {http://www.lrec-conf.org/proceedings/lrec2012/pdf/440_Paper.pdf} {The {I}celandic parsed historical corpus ({I}ce{P}a{HC})}.
\newblock In \emph{Proceedings of the Eighth International Conference on Language Resources and Evaluation ({LREC}'12)}, pages 1977--1984, Istanbul, Turkey. European Language Resources Association (ELRA).

\bibitem[{Rust et~al.(2021)Rust, Pfeiffer, Vuli{\'c}, Ruder, and Gurevych}]{rust-etal-2021-good}
Phillip Rust, Jonas Pfeiffer, Ivan Vuli{\'c}, Sebastian Ruder, and Iryna Gurevych. 2021.
\newblock \href {https://doi.org/10.18653/v1/2021.acl-long.243} {{How Good is Your Tokenizer? On the Monolingual Performance of Multilingual Language Models}}.
\newblock In \emph{Proceedings of the 59th Annual Meeting of the Association for Computational Linguistics and the 11th International Joint Conference on Natural Language Processing (Volume 1: Long Papers)}, pages 3118--3135, Online. Association for Computational Linguistics.

\bibitem[{Saleva and Lignos(2021)}]{saleva-lignos-2021-effectiveness}
Jonne Saleva and Constantine Lignos. 2021.
\newblock \href {https://doi.org/10.18653/v1/2021.eacl-srw.22} {The effectiveness of morphology-aware segmentation in low-resource neural machine translation}.
\newblock In \emph{Proceedings of the 16th Conference of the European Chapter of the Association for Computational Linguistics: Student Research Workshop}, pages 164--174, Online. Association for Computational Linguistics.

\bibitem[{Sandra(1994)}]{sandra1994morphology}
Dominiek Sandra. 1994.
\newblock The morphology of the mental lexicon: Internal word structure viewed from a psycholinguistic perspective.
\newblock \emph{Language and cognitive processes}, 9(3):227--269.

\bibitem[{Schmidt et~al.(2024)Schmidt, Reddy, Zhang, Alameddine, Uzan, Pinter, and Tanner}]{schmidt2024tokenization}
Craig~W Schmidt, Varshini Reddy, Haoran Zhang, Alec Alameddine, Omri Uzan, Yuval Pinter, and Chris Tanner. 2024.
\newblock \href {https://arxiv.org/abs/2402.18376} {Tokenization is more than compression}.
\newblock \emph{arXiv preprint arXiv:2402.18376}.

\bibitem[{Shin and You(2009)}]{shin2009hybrid}
Hyopil Shin and Hyunjo You. 2009.
\newblock Hybrid n-gram probability estimation in morphologically rich languages.
\newblock In \emph{Proceedings of the 23rd Pacific Asia Conference on Language, Information and Computation}, pages 511--520. Waseda University.

\bibitem[{Simov et~al.(2005)Simov, Osenova, Simov, and Kouylekov}]{SimOsSimKo2005}
Kiril Simov, Petya Osenova, Alexander Simov, and Milen Kouylekov. 2005.
\newblock Design and implementation of the bulgarian hpsg-based treebank.
\newblock \emph{Journal of Research on Language and Computation. Special Issue}, pages 495--522.

\bibitem[{Slobin(1973)}]{slobin1973cognitive}
Dan~I Slobin. 1973.
\newblock Cognitive prerequisites for the development of grammar.
\newblock In \emph{Studies of child language development}, pages 175--208. Holt, Rinehart, \& Winston.

\bibitem[{Slobin(2001)}]{slobin200114}
Dan~I Slobin. 2001.
\newblock {Form-function relations: how do children find out what they are?}
\newblock \emph{Language acquisition and conceptual development}, 3:406.

\bibitem[{Slobin(2013)}]{slobin2013crosslinguistic}
Dan~I Slobin. 2013.
\newblock Crosslinguistic evidence for the language-making capacity.
\newblock In \emph{The crosslinguistic study of language acquisition}, pages 1157--1256. Psychology Press.

\bibitem[{S{\o}gaard(2022)}]{sogaard-2022-ban}
Anders S{\o}gaard. 2022.
\newblock \href {https://doi.org/10.18653/v1/2022.emnlp-main.351} {Should we ban {E}nglish {NLP} for a year?}
\newblock In \emph{Proceedings of the 2022 Conference on Empirical Methods in Natural Language Processing}, pages 5254--5260, Abu Dhabi, United Arab Emirates. Association for Computational Linguistics.

\bibitem[{Song et~al.(2021)Song, Salcianu, Song, Dopson, and Zhou}]{song-etal-2021-fast}
Xinying Song, Alex Salcianu, Yang Song, Dave Dopson, and Denny Zhou. 2021.
\newblock \href {https://doi.org/10.18653/v1/2021.emnlp-main.160} {Fast {W}ord{P}iece {T}okenization}.
\newblock In \emph{Proceedings of the 2021 Conference on Empirical Methods in Natural Language Processing}, pages 2089--2103, Online and Punta Cana, Dominican Republic. Association for Computational Linguistics.

\bibitem[{Tan et~al.(2020)Tan, Joty, Varshney, and Kan}]{tan-etal-2020-mind}
Samson Tan, Shafiq Joty, Lav Varshney, and Min-Yen Kan. 2020.
\newblock \href {https://doi.org/10.18653/v1/2020.emnlp-main.455} {Mind your inflections! {I}mproving {NLP} for non-standard {E}nglishes with {B}ase-{I}nflection {E}ncoding}.
\newblock In \emph{Proceedings of the 2020 Conference on Empirical Methods in Natural Language Processing (EMNLP)}, pages 5647--5663, Online. Association for Computational Linguistics.

\bibitem[{Taul{\'e} et~al.(2008)Taul{\'e}, Mart{\'\i}, and Recasens}]{taule-etal-2008-ancora}
Mariona Taul{\'e}, M.~Ant{\`o}nia Mart{\'\i}, and Marta Recasens. 2008.
\newblock \href {http://www.lrec-conf.org/proceedings/lrec2008/pdf/35_paper.pdf} {{A}n{C}ora: Multilevel annotated corpora for {C}atalan and {S}panish}.
\newblock In \emph{Proceedings of the Sixth International Conference on Language Resources and Evaluation ({LREC}'08)}, Marrakech, Morocco. European Language Resources Association (ELRA).

\bibitem[{Tawfik et~al.(2019)Tawfik, Emam, Essam, Nabil, and Hassan}]{tawfik-etal-2019-morphology}
Ahmed Tawfik, Mahitab Emam, Khaled Essam, Robert Nabil, and Hany Hassan. 2019.
\newblock \href {https://doi.org/10.18653/v1/W19-4602} {Morphology-aware word-segmentation in dialectal {A}rabic adaptation of neural machine translation}.
\newblock In \emph{Proceedings of the Fourth Arabic Natural Language Processing Workshop}, pages 11--17, Florence, Italy. Association for Computational Linguistics.

\bibitem[{Touvron et~al.(2023)Touvron, Martin, Stone, Albert, Almahairi, Babaei, Bashlykov, Batra, Bhargava, Bhosale et~al.}]{touvron2023llama}
Hugo Touvron, Louis Martin, Kevin Stone, Peter Albert, Amjad Almahairi, Yasmine Babaei, Nikolay Bashlykov, Soumya Batra, Prajjwal Bhargava, Shruti Bhosale, et~al. 2023.
\newblock \href {https://arxiv.org/abs/2307.09288} {{Llama 2: Open foundation and fine-tuned chat models}}.
\newblock \emph{arXiv preprint arXiv:2307.09288}.

\bibitem[{Uzan et~al.(2024)Uzan, Schmidt, Tanner, and Pinter}]{uzan2024greed}
Omri Uzan, Craig~W Schmidt, Chris Tanner, and Yuval Pinter. 2024.
\newblock \href {https://arxiv.org/abs/2403.01289} {{Greed is All You Need: An Evaluation of Tokenizer Inference Methods}}.
\newblock \emph{arXiv preprint arXiv:2403.01289}.

\bibitem[{Vincze et~al.(2010)Vincze, Szauter, Almási, Móra, Alexin, and Csirik}]{vincze2010hungarian}
Veronika Vincze, Dóra Szauter, Attila Almási, György Móra, Zoltán Alexin, and János Csirik. 2010.
\newblock \href {http://www.lrec-conf.org/proceedings/lrec2010/pdf/465_Paper.pdf} {Hungarian dependency treebank}.
\newblock In \emph{Proceedings of the Seventh Conference on International Language Resources and Evaluation}, Valletta, Malta.

\bibitem[{Vylomova et~al.(2020)Vylomova, White, Salesky, Mielke, Wu, Ponti, Maudslay, Zmigrod, Valvoda, Toldova, Tyers, Klyachko, Yegorov, Krizhanovsky, Czarnowska, Nikkarinen, Krizhanovsky, Pimentel, Torroba~Hennigen, Kirov, Nicolai, Williams, Anastasopoulos, Cruz, Chodroff, Cotterell, Silfverberg, and Hulden}]{vylomova-etal-2020-sigmorphon}
Ekaterina Vylomova, Jennifer White, Elizabeth Salesky, Sabrina~J. Mielke, Shijie Wu, Edoardo~Maria Ponti, Rowan~Hall Maudslay, Ran Zmigrod, Josef Valvoda, Svetlana Toldova, Francis Tyers, Elena Klyachko, Ilya Yegorov, Natalia Krizhanovsky, Paula Czarnowska, Irene Nikkarinen, Andrew Krizhanovsky, Tiago Pimentel, Lucas Torroba~Hennigen, Christo Kirov, Garrett Nicolai, Adina Williams, Antonios Anastasopoulos, Hilaria Cruz, Eleanor Chodroff, Ryan Cotterell, Miikka Silfverberg, and Mans Hulden. 2020.
\newblock \href {https://doi.org/10.18653/v1/2020.sigmorphon-1.1} {{SIGMORPHON} 2020 shared task 0: Typologically diverse morphological inflection}.
\newblock In \emph{Proceedings of the 17th SIGMORPHON Workshop on Computational Research in Phonetics, Phonology, and Morphology}, pages 1--39, Online. Association for Computational Linguistics.

\bibitem[{Yamaguchi et~al.(2024)Yamaguchi, Villavicencio, and Aletras}]{yamaguchi2024empirical}
Atsuki Yamaguchi, Aline Villavicencio, and Nikolaos Aletras. 2024.
\newblock \href {https://arxiv.org/pdf/2402.10712} {An {E}mpirical {S}tudy on {C}ross-lingual {V}ocabulary {A}daptation for {E}fficient {G}enerative {LLM} {I}nference}.
\newblock \emph{arXiv preprint arXiv:2402.10712}.

\bibitem[{Yavrumyan and Danielyan(2020)}]{yavrumyan2020universal}
Marat~M. Yavrumyan and Anna~S. Danielyan. 2020.
\newblock {Universal Dependencies and the Armenian Treebank}.
\newblock \emph{Herald of the Social Sciences}, 2:231--244.

\bibitem[{Zeldes(2017)}]{Zeldes2017}
Amir Zeldes. 2017.
\newblock \href {https://doi.org/http://dx.doi.org/10.1007/s10579-016-9343-x} {The {GUM} corpus: Creating multilayer resources in the classroom}.
\newblock \emph{Language Resources and Evaluation}, 51(3):581--612.

\bibitem[{Zhou(2018)}]{zhou2018morphological}
Giulio Zhou. 2018.
\newblock Morphological zero-shot neural machine translation.

\bibitem[{Zouhar et~al.(2023)Zouhar, Meister, Gastaldi, Du, Sachan, and Cotterell}]{zouhar-etal-2023-tokenization}
Vil{\'e}m Zouhar, Clara Meister, Juan Gastaldi, Li~Du, Mrinmaya Sachan, and Ryan Cotterell. 2023.
\newblock \href {https://doi.org/10.18653/v1/2023.acl-long.284} {Tokenization and the noiseless channel}.
\newblock In \emph{Proceedings of the 61st Annual Meeting of the Association for Computational Linguistics (Volume 1: Long Papers)}, pages 5184--5207, Toronto, Canada. Association for Computational Linguistics.

\end{thebibliography}

\appendix

\section{MorphScore} \label{app:morphscore}

Table \ref{tab:morphscore_langs} reports the languages represented in MorphScore, their writing systems, language families, morphological types, and the number of items in each dataset.

\begin{table*}[h!]
  \centering
  \begin{tabular}{llllll}
    \hline
    \textbf{Language}      & \textbf{ISO}  & \textbf{Writing Sys.}  & \textbf{Lang. Family}   & \textbf{Morph. Type} & \textbf{Num. Items} \\
    &\textbf{639-3}& \textbf{(ISO 15924)} &&& \\
    \hline
    Armenian & hye &armn & Indo-European & agglutinative & 2000 \\
    Basque & eus &latn&Basque& agglutinative & 2000 \\
    Bulgarian & bul &cyrl&Indo-European& fusional &  2000 \\
    Cebuano & ceb &latn&Austronesian& agglutinative & 131 \\
    English & eng &latn&Indo-European&  fusional & 2000 \\
    Georgian & kat &geor&Kartvelian& agglutinative & 200 \\
    Greek & ell &grek&Indo-European& fusional & 112 \\
    Gujarati & guj &gujr&Indo-European& fusional & 547 \\
    Hungarian & hun &latn&Uralic& agglutinative & 2000 \\
    Icelandic & isl &latn&Indo-European& fusional & 1852 \\
    Indonesian & ind &latn&Austronesian& agglutinative & 1552\\
    Irish & gle &latn&Indo-European& fusional & 1877 \\
    Japanese & jpn &jpan&Japonic &agglutinative & 2000 \\
    Korean & kor &hang&Koreanic& agglutinative & 2000 \\
    Northern Kurdish & kmr &latn&Indo-European& fusional & 319 \\
    Persian & pes &arab&Indo-European& fusional & 2000 \\
    Slovenian & slv &latn&Indo-European& fusional & 2000 \\
    Spanish & spa &latn&Indo-European& fusional & 2000 \\
    Tamil & tam &taml&Dravidian& agglutinative & 884 \\
    Turkish & tur &latn&Turkic& agglutinative & 2000 \\
    Urdu & urd &arab&Indo-European& fusional & 1649 \\
    Zulu & zul &latn&Niger-Congo& agglutinative & 2000 \\
    \hline
  \end{tabular}
  \caption{Languages for which we created morphological datasets and evaluated MorphScore. }
  
  \label{tab:morphscore_langs}
\end{table*}

The sources used for each language are:
\begin{itemize}
    \item Bulgarian: UD\_Bulgarian-BTB train split \citep{SimOsSimKo2005} 
    \item English: UD\_English-GUM train split \citep{Zeldes2017} 
    \item Spanish: UD\_Spanish-AnCora train split \citep{taule-etal-2008-ancora} 
    \item Greek: UD\_Greek-GUD train split \citep{prokopidis-papageorgiou-2014-experiments} 
    \item Persian: UD\_Persian-PerDT train split \citep{rasooli-etal-2013-development} 
    \item Japanese: \citep{matsuzaki-etal-2024-j} 
    \item Korean: UD\_Korean-Kaist train split \citep{chun-etal-2018-building} 
    \item Turkish: UniMorph \citep{pimentel-ryskina-etal-2021-sigmorphon} 
    \item Indonesian: UD\_Indonesian-GSD \citep{larasati2011indonesian} 
    \item Hungarian: UD\_Hungarian-Szeged train split \citep{vincze2010hungarian}
    \item Urdu: UD\_Urdu-UDTB train split \citep{palmer2009hindi, bhathindi2017} 
    \item Slovenian: UD\_Slovenian-SSJ train split \citep{dobrovoljc-etal-2017-universal, dobrovoljc-ljubesic-2022-extending} 
    \item Tamil: UD\_Tamil-TTB train split \citep{tamil_ud}
    \item Georgian: UD\_Georgian-GLC test split \citep{lobzhanidze2022finite} 
    \item Armenian: UD\_Armenian-BSUT train split \citep{yavrumyan2020universal}
    \item Irish:  UD\_Irish-IDT train split \citep{lynn2016universal} 
    \item Icelandic: UD\_Icelandic-Modern train split \citep{rognvaldsson-etal-2012-icelandic}  
    \item Gujarati: UniMorph \citep{baxi2021morpheme} 
    \item Kurdish: UniMorph \citep{kirov-etal-2018-unimorph}  
    \item Cebuano: UD\_Cebuano-GJA test split \citep{ud_cebuano_gja}
    \item Basque: UD\_Basque-BDT train split \citep{aranzabe2015automatic} 
    \item Zulu: UniMorph \citep{vylomova-etal-2020-sigmorphon} 
\end{itemize}

Full MorphScore results for the tokenizers from \citet{chang2023multilinguality} are reported in Table \ref{tab:morphscore_com}.

\begin{table*}[htbp!]
\centering
\begin{tabular}{llll}
\hline
\textbf{Lang} & \textbf{Lang. Name} & \textbf{MorphScore} & \textbf{Morph. Type} \\ \hline
hye\_armn     & Armenian            & 0.634               & agg                  \\ 
eus\_latn     & Basque              & 0.724               & agg                  \\ 
bul\_cyrl     & Bulgarian           & 0.584               & fus                  \\ 
ceb\_latn     & Cebuano             & 0.806               & agg                  \\ 
eng\_latn     & English             & 0.781               & fus                  \\ 
kat\_geor     & Georgian            & 0.660               & agg                  \\ 
ell\_grek     & Greek               & 0.586               & fus                  \\
guj\_gujr     & Gujarati            & 0.347               & fus                  \\ 
hun\_latn     & Hungarian           & 0.739               & agg                  \\ 
isl\_latn     & Icelandic           & 0.574               & fus                  \\ 
ind\_latn     & Indonesian          & 0.708               & agg                  \\ 
gle\_latn     & Irish               & 0.468               & fus                  \\ 
jpn\_jpan     & Japanese            & 0.691               & agg                  \\ 
kor\_hang     & Korean              & 0.692               & agg                  \\ 
kmr\_latn     & Kurdish             & 0.202               & fus                  \\ 
pes\_arab     & Persian             & 0.345               & fus                  \\ 
slv\_latn     & Slovenian           & 0.650               & fus                  \\ 
spa\_latn     & Spanish             & 0.592               & fus                  \\ 
tam\_taml     & Tamil               & 0.435               & agg                  \\ 
tur\_latn     & Turkish             & 0.591               & agg                  \\ 
urd\_arab     & Urdu                & 0.747               & fus                  \\ 
zul\_latn     & Zulu                & 0.541               & agg                  \\ \hline
\end{tabular}
\caption{MorphScore results from Section \ref{sec:rq1}.}
\label{tab:morphscore_com}
\end{table*}

\end{document}